\documentclass{article}
\usepackage[square,numbers]{natbib}
\usepackage{authblk}
\usepackage{charter}
\usepackage{fullpage}
\usepackage{bbm}
\usepackage{appendix}
\usepackage{minitoc}

\usepackage{hyperref}       
\usepackage{url}            
\usepackage{booktabs}       
\usepackage{amsfonts}       
\usepackage{nicefrac}       
\usepackage{microtype}      
\usepackage[table]{xcolor}         
\usepackage{amsmath,amssymb,amsthm,amsfonts}
\usepackage{latexsym,bbm,graphicx,float,mathtools}
\usepackage{algorithmic}
\usepackage[export]{adjustbox}
\usepackage{capt-of}
\usepackage{booktabs}
\usepackage{parskip}

\usepackage[utf8]{inputenc}

\usepackage[noabbrev]{cleveref}
\usepackage{caption}
\usepackage{subcaption}
\usepackage{transparent}
\usepackage[inline]{enumitem}
\usepackage{multirow}
\usepackage{multicol}
\usepackage{algorithm}
\usepackage{xspace}

\usepackage{abstract}

\usepackage[subtle, mathdisplays=tight, charwidths=normal, leading=normal]{savetrees}

\newcommand{\name}{\textsc{Hallucination at a Glance:}\xspace}

\title{\textbf{\name Controlled Visual Editing and Fine-Grained Multimodal Learning}}

\author{
    Tianyi Bai$^*$\textsuperscript{1,2}, 
    Yuxuan Fan$^*$\textsuperscript{4}, 
    Jiantao Qiu$^\ddagger$\textsuperscript{2}, 
    Fupeng Sun\textsuperscript{6}, 
    Jiayi Song\textsuperscript{3},
    \vspace{-1em} \\
    Junlin Han\textsuperscript{5}, 
    Zichen Liu\textsuperscript{1}, 
    Conghui He$^{\dagger}$\textsuperscript{2}, 
    Wentao Zhang$^{\dagger}$\textsuperscript{3}, 
    Binhang Yuan$^{\dagger}$\textsuperscript{1}
    \vspace{0.4em} \\
    \textsuperscript{1}HKUST, \textsuperscript{2}Shanghai AI Lab, 
    \textsuperscript{3}Peking University, 
    \\
    \textsuperscript{4}HKUST(GZ), 
    \textsuperscript{5}Oxford University, 
    \textsuperscript{6}Imperial College London
    \vspace{0.4em}\\
    ${}^*$ Equal contribution.\ 
    ${}^\ddagger$ Project leader.\ 
    ${}^\dagger$ Corresponding authors.\\
    {wentao.zhang@pku.edu.cn, biyuan@ust.hk, qiujiantao@pjlab.org.cn}
    
}

\begin{document}
\maketitle

\begin{abstract}
\vspace{0.75em}
Multimodal large language models (MLLMs) have achieved strong performance on vision-language tasks but still struggle with fine-grained visual differences, leading to hallucinations or missed semantic shifts. We attribute this to limitations in both training data and learning objectives. To address these issues, we propose a controlled data generation pipeline that produces minimally edited image pairs with semantically aligned captions. Using this pipeline, we construct the Micro Edit Dataset (MED), containing over 50K image-text pairs spanning 11 fine-grained edit categories, including attribute, count, position, and object presence changes.
Building on MED, we introduce a supervised fine-tuning (SFT) framework with a feature-level consistency loss that promotes stable visual embeddings under small edits. We evaluate our approach on the Micro Edit Detection benchmark, which includes carefully balanced evaluation pairs designed to test sensitivity to subtle visual variations across the same edit categories.
Our method improves difference detection accuracy and reduces hallucinations compared to strong baselines, including GPT-4o. Moreover, it yields consistent gains on standard vision-language tasks such as image captioning and visual question answering. These results demonstrate the effectiveness of combining targeted data and alignment objectives for enhancing fine-grained visual reasoning in MLLMs.
Code and data are publicly released at {\url{https://github.com/beccabai/hallu_med}}.
\end{abstract}



\section{Introduction}
\begin{figure}[tb]
    \centering
    \includegraphics[width=0.9\textwidth]{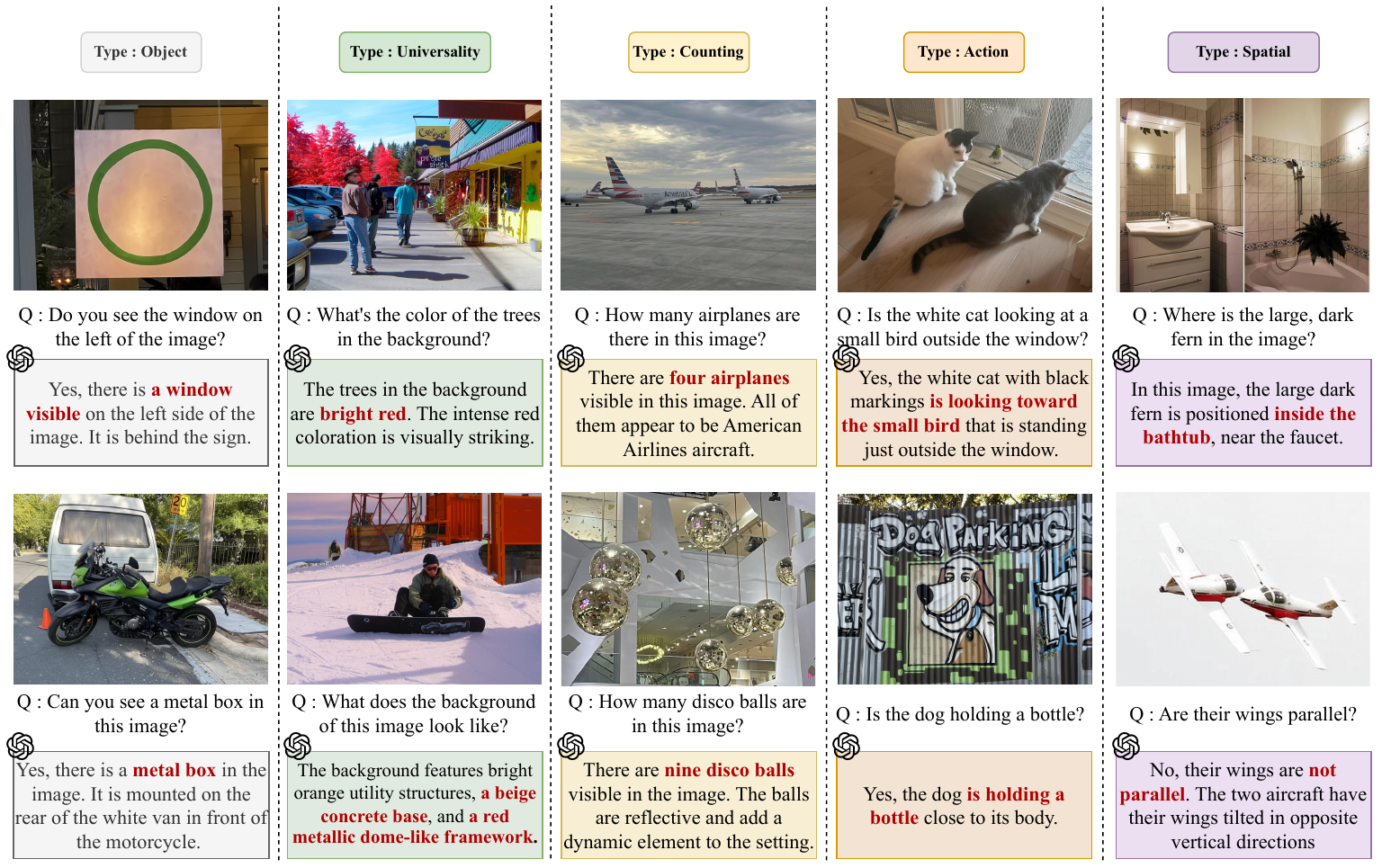}
    \vspace{-0.25cm}
    \caption{\textbf{GPT-4o exhibits hallucination errors on fine-grained visual question answering.} Instances are systematically selected to highlight failure cases in GPT-4o (Date accessed: May 08, 2025) visual question answering (VQA) performance across five hallucination-prone categories~\cite{fu2024tldr,lin2024evaluating}: Object, Universality, Counting, Action, and Spatial. Each row shows two examples where GPT-4o provides fluent yet inaccurate answers due to subtle misinterpretation of visual details. Hallucinated content is shown in red bold text, indicating model-generated descriptions that are not grounded in the image. All images are sourced from DOCCI and Visual Genome.}
    \label{fig:teaser}
    \vspace{-0.25cm}
\end{figure}

Multimodal large language models (MLLMs) have achieved impressive results on a wide range of vision-language tasks, including visual question answering, image captioning, and referring expression comprehension~\cite{caffagni2024revolution}. Despite this progress, such MLLMs sometimes become strikingly brittle when it comes to fine-grained visual understanding—the ability to detect and reason over small but semantically meaningful changes in images~\cite{peng2024synthesize}. As shown in Figure~\ref{fig:teaser}, even state-of-the-art models like GPT-4o frequently generate fluent but incorrect responses when faced with minimal edits involving object presence, count, spatial position, or attributes.

This limitation presents an obstacle for the deployment of MLLMs in real-world applications that demand high precision. Domains such as robotics~\cite{chen2024vlmimic}, industrial quality control~\cite{geipel2024towards}, medical imaging~\cite{panagoulias2024evaluating}, and assistive AI~\cite{yuan2024walkvlm} all require reliable grounding in subtle visual cues. Crucially, the distinctions that current models fail to capture—such as whether a tool is present, how many objects are in view, or the direction an object is facing—are trivial for humans, suggesting a fundamental mismatch between MLLM representations and the demands of fine-grained reasoning~\cite{bai2024hallucination,fu2024tldr}.

We attribute this deficiency to two intertwined factors: the lack of suitable training data and the limitations of current learning objectives. Large-scale web-crawled datasets rarely contain image pairs with tightly controlled, minimal semantic differences and aligned textual descriptions, making it difficult for models to learn how small visual changes map to linguistic shifts~\cite{zhang2025mllms} . Moreover, existing training paradigms do not explicitly enforce feature-level stability across such small perturbations, resulting in brittle visual-textual alignments that easily drift under fine-grained edits~\cite{mckinzie2024mm1}.

To address these issues, we propose a framework that combines targeted data construction with a new fine-grained alignment objective. We introduce a semantically controlled image editing pipeline that generates minimally modified image pairs with precise, contrastive captions. Using this pipeline, we construct the Micro Edit Dataset (MED)—a large-scale dataset comprising over 50K image-text pairs across 11 fine-grained edit types, including attribute changes, object insertion/removal, and so on.

To leverage this data, we design a supervised fine-tuning (SFT) strategy that incorporates a feature consistency regularization term, which encourages the image encoder to produce stable embeddings across visually similar inputs. This objective aligns visual representations more closely with semantic granularity, helping reduce hallucinations and improve robustness.

Finally, we introduce the Micro Edit Detection benchmark, a new evaluation suite designed to directly assess model sensitivity to subtle visual changes. Empirical results show that our method significantly outperforms strong baselines—including GPT-4o—on both edit detection and standard VQA and captioning tasks, demonstrating improved grounding and reduced hallucination.

Our contributions are as follows:
\begin{itemize}[topsep=5pt, leftmargin=*]
\vspace{-0.25em}
\item We introduce a semantically controlled image editing pipeline to generate high-quality, minimally different image-caption pairs at scale.
    \vspace{-0.25em}
    \item We construct and release the Micro Edit Dataset (MED) and the Micro Edit Detection benchmark, targeting fine-grained vision-language reasoning.
    \vspace{-0.25em}
    \item We propose a feature consistency regularization objective that improves representational stability under small semantic edits.
    \vspace{-0.25em}
    \item We demonstrate that our approach reduces hallucinations and improves fine-grained visual understanding across multiple benchmarks and models.
\end{itemize}
\vspace{-0.25cm}
\section{Related Work}
\vspace{-0.2cm}
\vspace{-0.2cm}
\textbf{Image editing.}
Image editing involves modifying the visual appearance, structure, or elements of an existing image \cite{huang2024diffusion}. 
While GANs \cite{goodfellow2020generative,karras2019style,pan2023drag} pioneered realistic image manipulation, recent diffusion models \cite{ho2020denoising,rombach2022high,song2020denoising} and flow-based models \cite{liu2022flow, lipman2022flow} have further advanced visual generation and editing capabilities.
To achieve more controllable and guided editing, various approaches have been developed, leveraging modalities such as textual instructions \cite{zhang2023magicbrush,brooks2023instructpix2pix,brack2024ledits}, masks \cite{huang2024smartedit,wang2021image}, layouts \cite{epstein2023diffusion,liu2023customizable}, segmentation maps \cite{matsunaga2022fine,yang2023imagebrush}, strokes \cite{meng2021sdedit,yang2023uni,liu2024magicquill}, references \cite{chen2024anydoor,chen2024unireal}, and point-dragging interfaces \cite{mou2023dragondiffusion,pan2023drag}. 
Recent state-of-the-art image editing models enable precise localized modifications while preserving the source content and maintaining consistency with the original distribution \cite{flux1filldev2024,liu2025step1x,gemini2025flash}. Leveraging these advances, we use the Gemini Flash 2.0 model \cite{gemini2025flash} to generate image pairs with subtle local differences, forming the foundation of our dataset.

\textbf{Multimodal LLMs.}
Multimodal large language models (MLLMs) integrate vision encoders with pretrained language models, enabling joint reasoning over text and images, as seen in \cite{wang2024qwen2, liu2023improved}. These typically combine visual encoders like CLIP \cite{radford2021learning} with adapter modules (MLPs, query-based transformers, or attention) to connect modalities. Discrete-token models such as Emu \cite{sun2023emu,sun2024generative,wang2024emu3} and Chameleon \cite{team2024chameleon} process all modalities as token sequences in a single transformer. Despite advancements, MLLMs still struggle with fine-grained visual understanding, often hallucinating details or missing subtle differences, especially in minimal-difference tasks \cite{li2024evaluating,fu2024tldr}. These challenges arise from datasets lacking controlled variability and misaligned training objectives. Our work seeks to address these issues by improving data quality and training methods to enhance fine-grained reasoning in MLLMs.

\textbf{Evaluating Multimodal LLMs}
The evaluation of Multimodal Large Language Models (MLLMs) has progressed from initial benchmarks like VQAv2 \cite{goyal2017making}, GQA \cite{hudson2019gqa}, and TextVQA \cite{singh2019towards} to newer frameworks such as MM-Vet \cite{yu2023mm, yu2024mm}, POPE \cite{li2023evaluating}, MMBench \cite{liu2024mmbench}, MMStar\cite{chen2024we} and MMVP \cite{zhang2024mmvp}, which focus on robustness, factual alignment, and hallucination. However, most benchmarks still emphasize coarse-grained tasks and overlook models’ struggles with subtle, semantically important visual differences—crucial for precision-sensitive use cases \cite{fu2024mme}. Fine-grained visual understanding remains a significant challenge, with MLLMs often missing minor but meaningful changes, resulting in hallucinations or errors. To address this, we introduce the Micro Edit Detection benchmark, which evaluates MLLMs on reasoning over minimally different image pairs, thereby complementing existing benchmarks and enabling more precise assessment of visual grounding and robustness.

\vspace{-0.25cm}

\section{The Micro Edit Dataset Construction}
\label{sec:med_construction}

\begin{figure}[tb]
    \centering
    \includegraphics[width=0.9\textwidth]{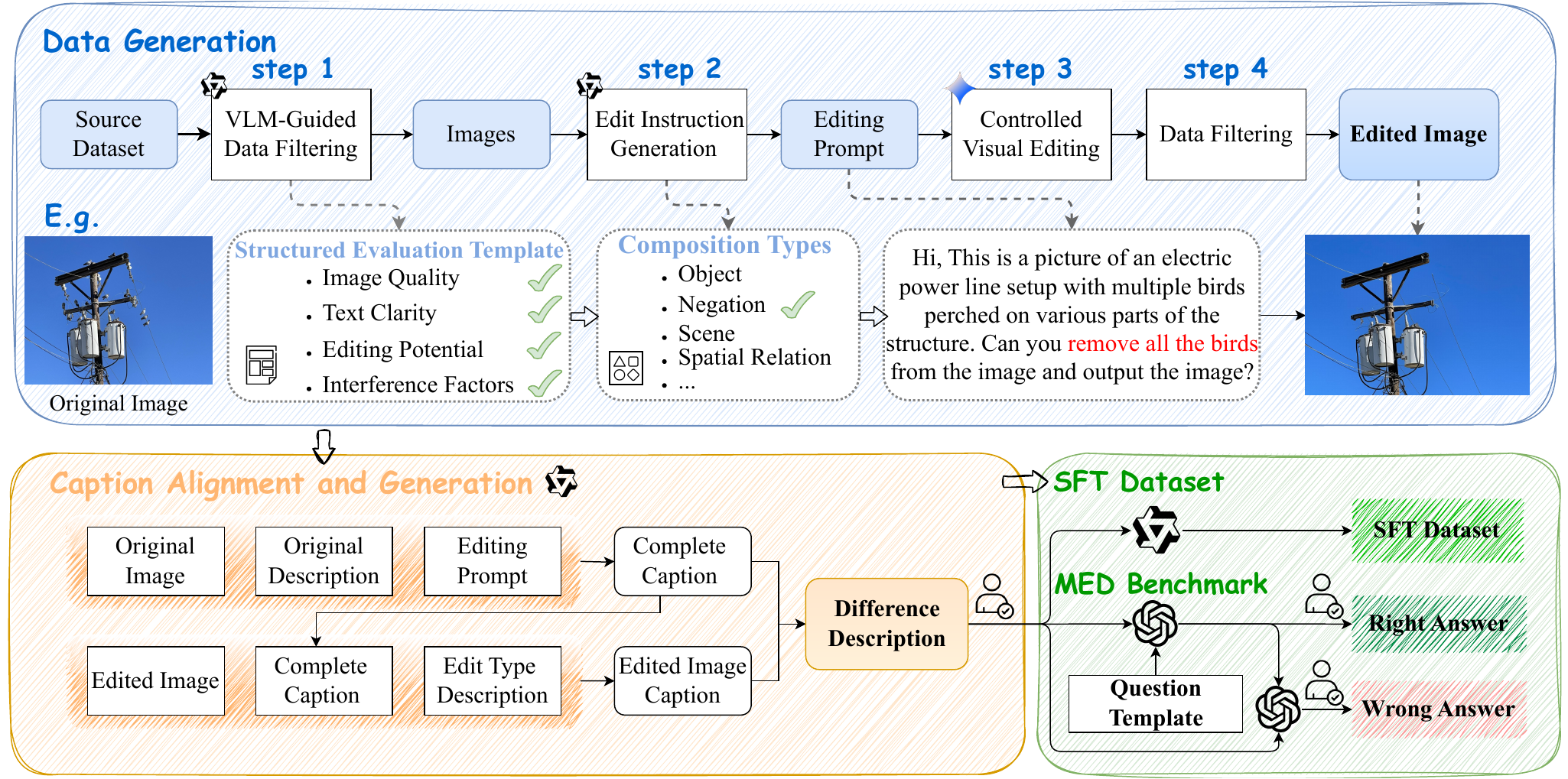}
    \vspace{-0.2cm}
    \caption{\textbf{Overview of the Micro Edit Dataset (MED) construction pipeline.} We begin with MLLM-guided filtering and controlled visual editing based on hallucination-prone change types. Caption alignment is performed via step-wise prompting to ensure consistency between original, edited images, and textual descriptions. The resulting data supports both supervised fine-tuning and benchmark evaluation for fine-grained visual reasoning.} 
    \label{fig:data_gen}
    \vspace{-0.5cm}
\end{figure}

To advance fine-grained visual understanding in MLLMs, we construct a dataset focused on subtle semantic differences, addressing gaps in existing benchmarks. Using DOCCI \cite{onoe2024docci} and Visual Genome \cite{krishna2017visual}, we develop a pipeline with filtering, semantic edit planning, and controlled image editing. In Section \ref{sec:data:gen}, we detail the complete pipeline used to generate the 50K-image Micro Edit Dataset, including the selection criteria, edit taxonomy, and caption alignment strategy. In Section \ref{sec:data:bench}, we describe how we construct the Micro Edit Detection (MED) benchmark from this dataset, which serves as a targeted evaluation tool for assessing fine-grained reasoning capabilities in MLLMs.

\vspace{-0.3cm}
\subsection{Data Generation}
\vspace{-0.25cm}
\label{sec:data:gen}
To construct a dataset for evaluating fine-grained visual understanding, we sample candidate image-text pairs from DOCCI and Visual Genome, which offer diverse, object-rich images and detailed captions—ideal for producing semantically minimal image pairs.

\textbf{MLLM-Guided Data Filtering.}
Due to quality and editability variations in raw image-caption pairs from DOCCI and Visual Genome, we employ a filtering pipeline using \texttt{Qwen-2.5-VL-72B}, a robust pretrained MLLM, for automatic evaluation. The model evaluates image quality, caption clarity, and editability using a structured template, considering factors such as sharpness, subject prominence, caption specificity, and whether minor edits would alter caption meaning. Only samples receiving a strict “Yes” are retained, ensuring alignment between visuals and text for semantic-preserving edits.

\begin{figure}[tb]
    \centering
    \includegraphics[width=1\textwidth]{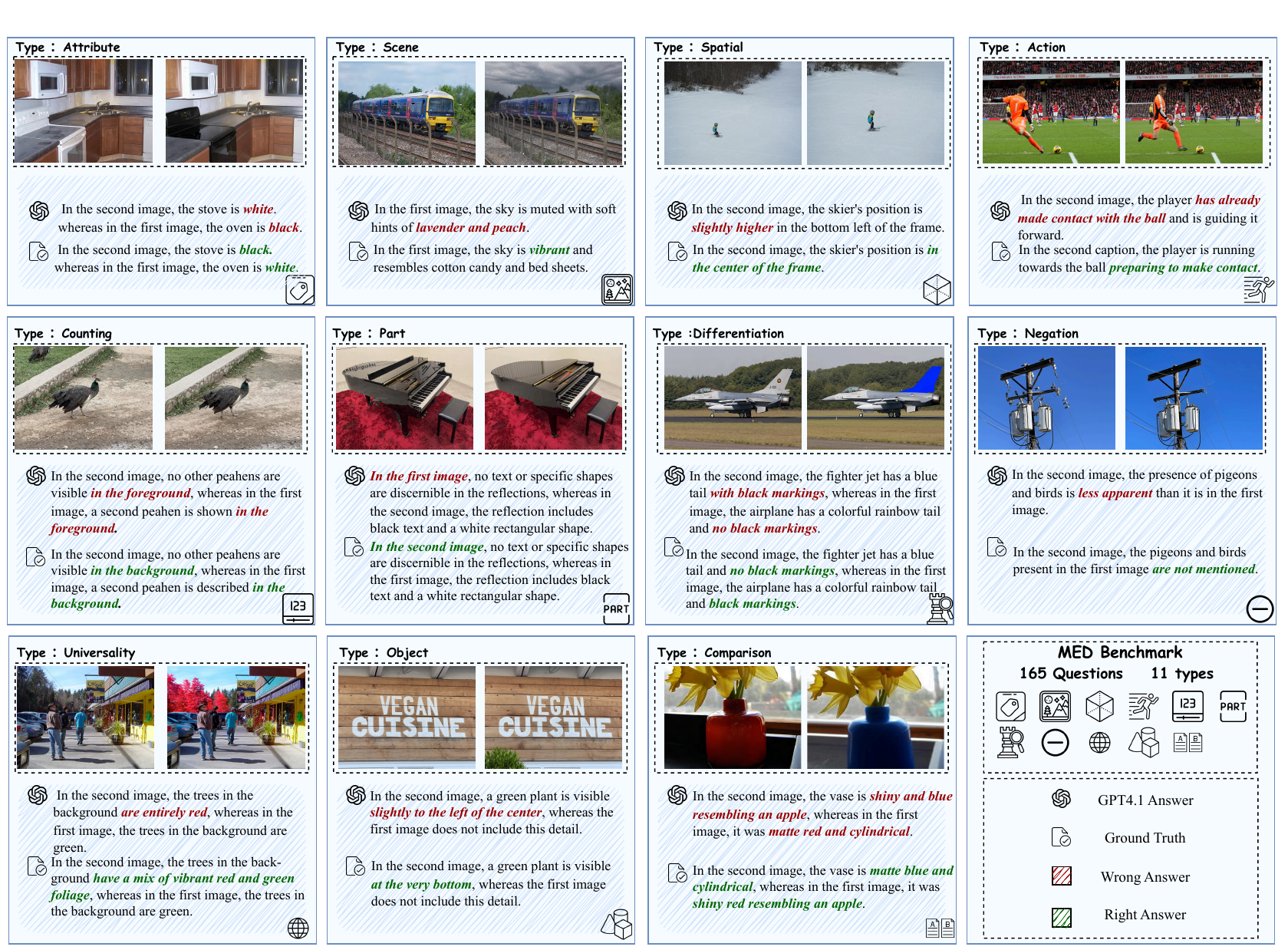}
    \vspace{-0.5cm}
    \caption{\textbf{Examples from the MED Benchmark covering all 11 composition types}. These categories reflect common sources of hallucination in MLLMs~\cite{fu2024tldr,lin2024evaluating} and include both basic changes (Object, Attribute, Scene, Spatial, Action, Part) and more complex transformations (Counting, Comparison, Differentiation, Negation, Universality). Each example shows an image pair, a difference question, and multiple-choice answers including the correct response and plausible distractors. }
    \label{fig:benchmark}
    \vspace{-0.5cm}
\end{figure}

\textbf{Controlled Visual Editing.}
After filtering, minimally different image pairs are generated using \texttt{Gemini Flash 2.0} for precise edits, guided by edit instructions created by \texttt{Qwen-2.5-VL-72B}. These instructions are based on eleven compositional change categories linked to common MLLM hallucination sources \citep{lin2024evaluating, fu2024tldr}, such as object manipulation, attribute changes, spatial reconfiguration, and complex edits like counting and comparison (see Figure~\ref{fig:benchmark}). For each image, the most suitable category—favoring non-object edits—is chosen, and a fixed-format natural language prompt is generated. These instructions ensure minimal yet meaningful changes, resulting in contrastive image pairs with aligned original and updated captions reflecting the visual edits.

\textbf{Data Filtering.} 
To ensure the generated image pairs differ only in subtle ways, we introduce a visual similarity filtering stage inspired by \citep{tong2024eyes}. Using CLIP embeddings from the \texttt{clip-vit-base-patch32} model, we calculate cosine similarity and discard pairs below a 0.7 threshold following the MMVP \citep{zhang2024mmvp}. This filtering retains pairs with subtle, meaningful edits, emphasizing minimal visual differences and supporting fine-grained visual understanding evaluation.

\vspace{-0.5em}
\subsection{Caption Alignment and Generation}
\label{sec:data:gen}
\vspace{-0.5em}

To ensure image pairs have semantically precise and visually grounded captions, we design a structured caption generation pipeline shown in~\ref{fig:data_gen}. This pipeline both verifies caption correctness after editing and removes inconsistent image-text pairs. Although automatic fine-grained alignment is difficult, we leverage state-of-the-art MLLMs for captioning and text comparison. Specifically, we break the alignment check into four steps, each handled independently by \texttt{Qwen2.5-VL-72B}.

\textbf{Step 1: Caption Completion for Original Image.}
Original captions in DOCCI and Visual Genome are often brief and may omit elements relevant to the edit prompt. To create a faithful reference, we prompt the model with the original image, its caption, and the edit prompt, asking it to revise the caption only if the image visibly includes the specified elements. This ensures the original caption is comprehensive yet strictly grounded in the visible content.

\textbf{Step 2: Caption Generation for Edited Image.}
With the edited image in hand, we generate a new caption that accurately reflects the updated content. The model is given the complete original caption as an example, and a description of the editing type (e.g., “change in spatial relation”) to focus its attention on the most relevant aspects.

\textbf{Step 3: Generating Difference Description.}
For contrastive supervision, we compare captions from Step 1 and Step 2, then use \texttt{Qwen3-32B}, a strong text-only LLM, to generate a concise description of their key difference. The model identifies the most salient change and classifies it into a predefined semantic edit type. This difference sentence is later used during training or evaluation to support fine-grained reasoning over minimal edits.

\textbf{Step 4: Human Verification and Iterative Refinement.}
To validate data quality, we manually inspect 1,000 sampled image pairs across all eleven visual change types. Annotators evaluate whether captions and difference descriptions accurately align with the visual content. Based on issues found—like under-specified changes or inaccurate attributes—we refine prompts and reprocess affected samples using Qwen-VL-Max. This helps eliminate systematic errors and ensures high data quality. To enhance difference descriptions, \texttt{Qwen3-32B} and \texttt{Qwen2.5-VL-72B} rewrite each sample with richer vocabulary and improved clarity. Visual changes are rephrased into a QA format, forming MED, an instruction dataset with over 50K high-quality samples paired with the images.

\vspace{-0.5em}
\subsection{The Micro Edit Detection (MED) Benchmark}
\label{sec:data:bench}
\vspace{-0.5em}

Using MED, we construct a benchmark for evaluating fine-grained visual reasoning in multimodal models. We select image pairs with high semantic similarity (CLIP cosine similarity > 0.95), and for each pair, generate nine rephrased questions (see Appendix~\ref{sec:appendix:benchmark_examples}). \texttt{GPT-4o} produces the correct answer and three subtle distractors for each question. All question-answer sets are manually verified and categorized into eleven edit types. To prevent contamination, benchmark samples are excluded from the fine-tuning dataset. The final benchmark contains 165 questions evenly distributed across all edit types, offering a robust tool for fine-grained visual reasoning evaluation.

To address potential concerns about overfitting to synthetic data, as both the training set and benchmark stem from the same controlled editing pipeline, we augment the evaluation with real-world image pairs. The MED-Real Set is created by sampling 35 minimally different image pairs from the MMVP benchmark \cite{tong2024eyes}, selecting those with CLIP similarity scores above 0.95 to capture subtle yet meaningful visual variations. Question-answer items are generated for each pair using the same fine-grained reasoning format as in the main benchmark. This expands the evaluation set to 200 items, combining 165 synthetic edit pairs and 35 real-world pairs, offering a more comprehensive assessment of sensitivity to controlled differences and real-world generalization.

\vspace{-0.3cm}
\section{Fine-Grained Multimodal
Learning for MLLMs}
\vspace{-0.2cm}
In this section, we present our approach for enhancing fine-grained visual difference understanding in MLLMs. Section~\ref{sec:method:setup} formalizes the problem setting. Section~\ref{sec:method:downstream} analyzes the impact of semantic misalignment and objective mismatch on downstream performance. Finally, Section~\ref{sec:method:sft} introduces a supervised fine-tuning (SFT) strategy that leverages semantically controlled data augmentation to improve the MLLM’s capacity for fine-grained multimodal reasoning.

\vspace{-0.2cm}

\subsection{Problem Setup}
\vspace{-0.2cm}

\label{sec:method:setup}
We formalize the task of fine-grained visual difference detection using MLLMs. Let $\mathcal{X}$ denote the space of images, equipped with a perceptual premetric $d_{\mathcal{X}}$ (e.g., Learned Perceptual Image Patch Similarity (LPIPS)). Let $\mathcal{T}$ denote the space of natural language captions, equipped with a semantic premetric $d_{\mathcal{T}}$ (e.g., edit distance or BERT-score). Define the ground truth difference operator $\Delta_{\mathcal{T}}(t_1, t_2)$ as the natural language description of the difference between two captions $t_1$ and $t_2$. 
A MLLM $\texttt{M}_{\theta} = \{\texttt{I}_{\theta}, \texttt{T}_{\theta}, \texttt{Z}_{\theta}\}$ consists of an image encoder $\texttt{I}_{\theta}$, a text encoder $\texttt{T}_{\theta}$, and an text decoder $\texttt{Z}_{\theta}$, with parameters $\theta$.
We define two loss functions: the autoregressive captioning loss $l_{\text{cap}}$ and a CLIP-style contrastive loss $l_{\text{clip}}$. Given a distribution $\mathcal{P}(\mathcal{X})$ and a ground-truth captioning function $f: \mathcal{X} \to \mathcal{T}$, MLLM training aims to minimize the population risk (equation (4) in \cite{yu2022coca}):
\begin{align}\label{eq:ex_loss}
\begin{small}
R(\theta) = \mathbb{E}_{x\sim \mathcal{P}(\mathcal{X}),\ t = f(x)} \left[
\lambda_{\text{cap}} \cdot l_{\text{cap}}(\texttt{Z}_{\theta}[\texttt{I}_{\theta}(x)], t) + \lambda_{\text{clip}} \cdot l_{\text{clip}}(\texttt{I}_{\theta}(x) - \texttt{T}_{\theta}(t))
\right],
\end{small}
\end{align}
where $\lambda_{\text{cap}}$ and $\lambda_{\text{clip}} > 0$ are penalty parameters.

The empirical risk over a dataset $\mathcal{D} = \{(x_i, t_i)\}_{i=1}^N$ is:
\begin{align}\label{eq:em_loss}
\begin{small}
\hat{R}(\theta) = \frac{1}{|\mathcal{D}|} \sum_{(x_i, t_i) \in \mathcal{D}} \left[
\lambda_{\text{cap}} \cdot l_{\text{cap}}(\texttt{Z}_{\theta}[\texttt{I}_{\theta}(x_i)], t_i) + \lambda_{\text{clip}} \cdot l_{\text{clip}}(\texttt{I}_{\theta}(x_i) - \texttt{T}_{\theta}(t_i))
\right].
\end{small}
\end{align}

\vspace{-0.2cm}
\subsection{Downstream Image Difference Task}
\vspace{-0.2cm}

\label{sec:method:downstream}
We consider downstream tasks where the goal is to describe differences between two visually similar images \( x_1, x_2 \in \mathcal{X} \), satisfying \( d_{\mathcal{X}}(x_1, x_2) \leq \epsilon \). The performance of a MLLM on this task can be evaluated using the generalization error:
\begin{equation}
\begin{small}
G(\theta) = \mathbb{E}_{x_1, x_2 \sim \mathcal{P}(\mathcal{X}),\ t_1 = f(x_1),\ t_2 = f(x_2)}
\left[
l_{\text{cap}}\left(\texttt{Z}_{\theta}[\texttt{I}_{\theta}(x_1) - \texttt{I}_{\theta}(x_2)],\ \Delta_{\mathcal{T}}(t_1, t_2)\right)
\right],
\end{small}
\end{equation}

where $\texttt{Z}_{\theta}[\texttt{I}_{\theta}(x_1) - \texttt{I}_{\theta}(x_2)]$ means that translating the feature-level image difference into a language-level description. Ideally, we would like the empirical minimizer \( \hat{\theta} \in \arg\min \hat{R}(\theta) \) to achieve low generalization error, i.e., \( G(\hat{\theta}) \) is small. However, even state-of-the-art MLLMs (e.g., GPT-4o) still underperform on fine-grained difference description tasks. This raises a central question: \emph{why does this performance gap persist}?
There are two primary contributing factors. \underline{First}, there may be \emph{semantic under-specification}, where captions omit salient visual details, or \emph{caption incompleteness}, where not all semantically meaningful aspects of an image \( x \) are described in the corresponding caption \( t \). More broadly, such cases fall under \emph{visual-textual misalignment}, where the semantic content of the image and caption diverge. As these misalignments can range from mild to severe, we model them as noisy captions \( \tilde{t} = \tilde{f}(x) \), where \( \tilde{f} \) is an arbitrary, potentially biased captioning function.
\underline{Second}, even if the MLLM is trained on a clean dataset, a generalization gap can still arise when the model is trained to minimize \( R(\theta) \) but evaluated on the downstream objective \( G(\theta) \). For example, models in prior works \cite{bai2025qwen2, grattafiori2024llama, liu2023improved, wang2024qwen2} were trained on datasets with rewritten captions. However, our experiments show that these models still perform poorly on tasks that require sensitivity to visual differences. The underlying intuition is that standard MLLMs trained using the loss in equation~\eqref{eq:em_loss} lack the architectural capacity to explicitly recognize and describe fine-grained differences between similar images.

\vspace{-0.2cm}
\subsection{Supervised Fine-tuning}
\label{sec:method:sft}
\vspace{-0.2cm}

Motivated by the preceding two factors, we assume the MLLM \( \texttt{M}_{\theta} \) is trained on a \emph{noisy dataset} \( \mathcal{D}_\eta \), where each image-caption pair \( (x, t) \) satisfies that \( t = f(x) \) with probability \( \eta \), and \( t = \tilde{f}(x) \) otherwise. That is, with probability \( \eta \), the caption faithfully describes the image; with probability \( 1 - \eta \), it reflects an imperfect or incomplete description.

Accordingly, the baseline MLLM \( \texttt{M}_\theta \) is trained on \( \mathcal{D}_\eta \) by minimizing the empirical risk:
\vspace{-0.2cm}
\begin{align}\label{eq:em_corrupted_loss}
\begin{small}
\hat{R}_{\eta}(\theta) = \frac{1}{|\mathcal{D}_{\eta}|} \sum_{(x_i, t_i) \in \mathcal{D}_{\eta}} \left[
\lambda_{\text{cap}} \cdot l_{\text{cap}}(\texttt{Z}_{\theta}[\texttt{I}_{\theta}(x_i)], t_i) 
+ \lambda_{\text{clip}} \cdot l_{\text{clip}}(\texttt{I}_{\theta}(x_i) - \texttt{T}_{\theta}(t_i))
\right].
\end{small}
\end{align}
\vspace{-0.2cm}

This objective corresponds to the population-level risk:
\begin{align}\label{eq:ex_corrupted_loss}
R_{\eta}(\theta) = \eta \cdot R(\theta) + (1 - \eta) \cdot \tilde{R}(\theta),
\end{align}
where
\vspace{-0.2cm}
\begin{equation}
\begin{small}
    \tilde{R}(\theta) = \mathbb{E}_{x \sim \mathcal{P}(\mathcal{X}),\ t = \tilde{f}(x)} \left[
\lambda_{\text{cap}} \cdot l_{\text{cap}}(\texttt{Z}_{\theta}[\texttt{I}_{\theta}(x)], t) 
+ \lambda_{\text{clip}} \cdot l_{\text{clip}}(\texttt{I}_{\theta}(x) - \texttt{T}_{\theta}(t))
\right].
\end{small}
\end{equation}
\vspace{-0.2cm}

\begin{table*}[t]
\scriptsize
\centering
\newcolumntype{C}{>{\centering\arraybackslash}p{6.8mm}}
\resizebox{\textwidth}{!}{
\begin{tabular}{lCCCCCCCCCCCC}
\toprule
\textbf{Model} & Object & Attr. & Scene & Spatial & Action & Part & Count & Differ. & Compar. & Neg. & Univ. & Avg \\
\midrule

Human & 85.71 & 100 & 100 & 100 & 100 & 93.75 & 100 & 75 & 100 & 100 & 92.86 & \textbf{95.21} \\
\midrule
\multicolumn{13}{c}{\textbf{API-based Models}} \\
\midrule
GPT-4o-2024-08-06 & 57.14 & 56.25 &61.54 & 57.14 & 33.33 & 43.75 & 50.00 & \textbf{66.67}  & 31.58 & 42.86 & \textbf{64.29} & 51.32 \\
GPT-4.1-2025-04-14 & \textbf{71.43} & \textbf{62.50} & 61.54 & \textbf{50.00} & 33.33 & \textbf{50.00} & \textbf{61.11} &\textbf{66.67}  & 47.37 & 42.86 & 50.00 & \textbf{54.26} \\
Claude3.7 Sonnet & 57.14 & 43.75 & 46.15 & 35.71 & 46.67 & 25.00 & 38.89 & 50.00 & 15.79 & 42.86 & 42.86 & 40.44 \\
Qwen-VL-Plus-25-03 & 57.14 & 56.25 &61.54& 42.86 & 33.33 & 31.25 & 38.89 & \textbf{66.67} & 47.37 & \textbf{50.00} & 35.71 & 47.36 \\
Doubao-1.5-vision-pro & 69.23 & 37.50 & \textbf{69.23} & \textbf{50.00} & \textbf{50.00} & \textbf{50.00} & 38.89 & \textbf{66.67} & \textbf{52.63} & 42.86 & 42.86 & 51.81 \\
\midrule
\multicolumn{13}{c}{\textbf{Open-source Models}} \\
\midrule
Qwen2-VL-7B & 35.71 & 43.75 & \textbf{61.54}  & 42.86 & 46.67 & 31.25 & 50.00 & 33.33 & 21.05 & 28.57 & 28.57 & 38.48 \\
Qwen2-VL-7B (ours) 
& \cellcolor{green!10}42.86 
& \cellcolor{green!10}43.75 
& \cellcolor{red!5}53.85 
& \cellcolor{green!10}\textbf{57.14} 
& \cellcolor{green!10}\textbf{53.33} 
& \cellcolor{green!10}\textbf{50.00} 
& \cellcolor{green!10}\textbf{55.56} 
& \cellcolor{green!10}\textbf{58.33}
& \cellcolor{green!10}36.84 
& \cellcolor{green!10}42.86
& \cellcolor{green!10}28.57 
& \cellcolor{green!10}47.55 \\
Qwen2.5-VL-7B & 53.85 & 50.00 & 38.46 & 42.86 & 12.50 & 18.75 & 44.44 & 50.00 & 26.32 & 42.86 & 57.14 & 39.74 \\
Qwen2.5-VL-7B (ours) 
& \cellcolor{green!10}57.14 
& \cellcolor{green!10}56.25 
& \cellcolor{green!10}53.84
& \cellcolor{green!10}42.86
& \cellcolor{green!10}46.67 
& \cellcolor{green!10}43.75
& \cellcolor{green!10}\textbf{55.56}
& \cellcolor{green!10}50.00 
& \cellcolor{green!10}\textbf{47.37}
& \cellcolor{green!10}\textbf{50.00}
& \cellcolor{green!10}\textbf{64.29} 
& \cellcolor{green!10}\textbf{51.61} \\
LLaVA-V1.6-7B & \textbf{64.29} & 37.50 & 30.77 & 21.43 & 33.33 & 25.00 & 27.78 & 25.00 & 26.32 & 21.43 & 28.57 & 31.04 \\
LLaVA-V1.6-7B (ours) 
& \cellcolor{red!5}57.14 
& \cellcolor{green!10}37.50 
& \cellcolor{green!10}46.15 
& \cellcolor{green!10}42.86 
& \cellcolor{green!10}46.67 
& \cellcolor{green!10}37.50 
& \cellcolor{green!10}44.44 
& \cellcolor{green!10}33.33 
& \cellcolor{green!10}42.11 
& \cellcolor{green!10}28.57 
& \cellcolor{green!10}28.57 
& \cellcolor{green!10}40.44 \\
LLaMA-3.2-11B & 46.15 & 43.75 & 38.46 & 50.00 & 50.00 & 25.00 & 22.22 & 33.33 & 15.79 & 21.43 & 35.71 & 34.71 \\
LLaMA-3.2-11B (ours) 
& \cellcolor{green!10}38.46 
& \cellcolor{green!10}\textbf{62.50} 
& \cellcolor{green!10}46.15 
& \cellcolor{green!10}35.71 
& \cellcolor{red!5}37.50 
& \cellcolor{green!10}37.50 
& \cellcolor{green!10}33.33 
& \cellcolor{green!10}41.67 
& \cellcolor{green!10}31.58 
& \cellcolor{green!10}42.86
& \cellcolor{green!10}42.86 
& \cellcolor{green!10}40.92  \\
\bottomrule
\end{tabular}
}
\caption{
\textbf{Model performance on the MED benchmark subtasks.} 
Each column corresponds to a specific type of task in MED benchmark: Object, Attr. (Attribute), Scene, Spatial, Action, Part, Count, Differ. (Differentiation), Compar. (Comparison), Neg. (Negation), Univ. (Universality), and Avg (overall accuracy).
Bold values indicate the best performance per column within each model category.
We use colors to highlight whether model trained with our method \colorbox{green!10}{increases} or \colorbox{red!10}{decreases} performance compared to the original model.
}
\vspace{-1.5em}
\label{tab:med_results}

\end{table*}

Let \( \hat{\theta}_\eta \) and \( \theta^*_\eta \) denote the empirical and population risk minimizers under noisy supervision. Due to the presence of label noise and a mismatch between the training and evaluation objectives, the downstream generalization error \( G(\hat{\theta}_\eta) \) may remain large. As a result, the model \( \texttt{M}_{\hat{\theta}_\eta} \) exhibits suboptimal performance on fine-grained image difference tasks.

To mitigate this issue, we propose a supervised fine-tuning (SFT) approach using a curated dataset focused on image differences. Given a noisy dataset \( \mathcal{D}_\eta \), for each pair \( (x_i, \tilde{t}_i) \), we construct an augmented triplet as follows.

\vspace{-0.5em}
\begin{itemize}[topsep=5pt, leftmargin=*]
    \item Create the clean caption \( t_i = f(x_i) \);
    \item Edit \( \tilde{t}_i \) to obtain \( \hat{t}_i \), and use an image generator (e.g., Gemini) to produce a visually similar image \( \hat{x}_i \) such that \( d_{\mathcal{X}}(x_i, \hat{x}_i) \leq \epsilon \) and \( \hat{t}_i = f(\hat{x}_i) \);
    \item Use a prompting engine \( \texttt{S}_{\phi} \) (e.g., an LLM) to generate a fine-grained natural language difference description \( \texttt{S}_{\phi}(t_i, \hat{t}_i) \), where \( \phi \) represents all parameters of the generator.
\end{itemize}

This process yields a curated dataset \( \mathcal{D}_{\text{edit}} \) for post-training. We fine-tune the model \( \texttt{M}_{\hat{\theta}_\eta^*} \) on this dataset by minimizing the following empirical SFT loss:
\begin{align}\label{eq:SFT_loss}
\begin{small}
\hat{R}_{\text{SFT}}(\theta) = \frac{1}{|\mathcal{D}_{\text{edit}}|} \sum_{(x_i, \hat{x}_i, \texttt{S}_{\phi}(t_i, \hat{t}_i)) \in \mathcal{D}_{\text{edit}}} \left[
l_{\text{cap}}\left(\texttt{Z}_{\theta}[\texttt{I}_{\theta}(x_i) - \texttt{I}_{\theta}(\hat{x}_i)],\ \texttt{S}_{\phi}(t_i, \hat{t}_i)\right)
\right].
\vspace{-0.25cm}
\end{small}
\end{align}
By minimizing \( \hat{R}_{\text{SFT}}(\theta) \), the model parameter is updated from the noisy initialization \( \hat{\theta}_\eta \) to a refined estimate \( \hat{\theta}_{\text{SFT}} \). Empirically, we find that the post-trained model $\texttt{M}_{\hat{\theta}_{\text{SFT}}} $ achieves significantly better performance on visual difference description tasks than the baseline $ \texttt{M}_{\hat{\theta}_\eta} $. Moreover, the enhanced ability to reason about fine-grained visual changes also improves performance on classical downstream tasks, as it fosters deeper visual understanding and architectural awareness of subtle semantic shifts between images.

\begin{table}[t]
\scriptsize
\centering
\newcolumntype{C}{>{\centering\arraybackslash}p{9mm}}
\resizebox{\columnwidth}{!}{
\begin{tabular}{lCCCCCCCCC}
\toprule
\textbf{Model} & \textbf{Pope} & \textbf{Coarse} & \textbf{Fine} & \textbf{Visual\_Sim} & \textbf{Visual\_Corr} & \textbf{Count} & \textbf{MMVP} & \textbf{Ave} & \textbf{MME} \\
\midrule
Qwen2-VL-7B                 & 92.50 & 71.21 & 48.24 & 51.11 & 30.23 & 55.83 & 31.33 & 54.35 & 1679.52 \\
Qwen2-VL-7B (Ours)          
& \cellcolor{green!10}96.27 
& \cellcolor{green!10}73.92 
& \cellcolor{red!5}46.16 
& \cellcolor{green!10}51.85 
& \cellcolor{green!10}33.72 
& \cellcolor{green!10}59.17 
& \cellcolor{green!10}32.67 
& \cellcolor{green!10}56.25 
& \cellcolor{green!10}1681.27 \\
Qwen2.5-VL-7B               & 96.29 & 73.95 & 57.35 & 49.63 & 33.72 & 50.00 & 27.33 & 55.47 & 1685.14 \\
Qwen2.5-VL-7B (Ours)        
& \cellcolor{green!10}97.52 
& \cellcolor{green!10}75.97 
& \cellcolor{green!10}59.36 
& \cellcolor{green!10}51.85 
& \cellcolor{green!10}37.79 
& \cellcolor{green!10}59.17 
& \cellcolor{green!10}28.00 
& \cellcolor{green!10}58.52 
& \cellcolor{green!10}1701.87 \\
LLaVA-V1.6-7B            & 95.56 & 58.28 & 31.93 & 51.11 & 21.51 & 45.83 & 28.67 & 47.56 & 1441.89 \\
LLaVA-V1.6-7B (Ours)     
& \cellcolor{green!10}97.39 
& \cellcolor{red!5}56.74 
& \cellcolor{green!10}35.13 
& \cellcolor{red!5}48.14 
& \cellcolor{green!10}24.42 
& \cellcolor{green!10}49.17 
& \cellcolor{green!10}30.00 
& \cellcolor{green!10}48.71 
& \cellcolor{red!5}1420.57 \\
LLaMA-3.2-11B               & --    & 69.03 & 48.94 & 43.70 & 20.93 & 44.17 & 26.00 & 42.13 & 1421.71 \\
LLaMA-3.2-11B (Ours)        
& --    
& \cellcolor{green!10}72.60 
& \cellcolor{red!5}47.21 
& \cellcolor{green!10}45.93 
& \cellcolor{red!5}19.19 
& \cellcolor{green!10}50.00 
& \cellcolor{green!10}28.00 
& \cellcolor{green!10}43.82 
& \cellcolor{green!10}1430.67 \\
\bottomrule
\end{tabular}}
\caption{
\textbf{Performance of models on other benchmarks.} 
\colorbox{green!10}{Green} indicates improved performance, and \colorbox{red!5}{red} indicates decreased performance after fine-tuning. 
Coarse and Fine correspond to the \textit{Coarse Perception} and \textit{Fine-grained Perception} sub-tasks in the MMStar; Visual\_Sim, Visual\_Corr, and Count correspond to the \textit{Visual\_Similarity}, \textit{Visual\_Correspondence}, and \textit{Counting} sub-tasks in BLINK, respectively; Pope reports POPE precision, and MME is the MME perception score.
}
\vspace{-0.5cm}
\label{tab:general_bench_results}
\end{table}
\begin{table}[t]
\scriptsize
\centering
\newcolumntype{C}{>{\centering\arraybackslash}p{12mm}} 
\resizebox{\columnwidth}{!}{
\begin{tabular}{lCCCCCCCCC}
\toprule
\textbf{Model} & \textbf{Pope} & \textbf{Coarse} & \textbf{Fine} & \textbf{Visual\_Sim} & \textbf{Visual\_Corr} & \textbf{Count} & \textbf{MMVP} & \textbf{Ave} & \textbf{MME} \\
\midrule
Qwen2-VL-7B                       & 92.50 & 71.21 & 48.24 & 51.11 & 30.23 & 55.83 & 31.33 & 54.35     & 1679.52 \\
Qwen2-VL-7B + trained-ViT        
& \cellcolor{green!10}93.79 
& \cellcolor{green!10}73.81 
& \cellcolor{green!10}49.17 
& \cellcolor{yellow!10}51.11 
& \cellcolor{green!10}31.40 
& \cellcolor{red!5}53.28 
& \cellcolor{green!10}32.00 
& \cellcolor{green!10}54.94 
& \cellcolor{red!5}1668.18 \\
Qwen2-VL-7B (Ours)          
& \cellcolor{green!10}96.27 
& \cellcolor{green!10}73.92 
& \cellcolor{red!5}46.16 
& \cellcolor{green!10}51.85 
& \cellcolor{green!10}33.72 
& \cellcolor{green!10}59.17 
& \cellcolor{green!10}32.67 
& \cellcolor{green!10}56.25 
& \cellcolor{green!10}1681.27 \\
\midrule
Qwen2.5-VL-7B                     & 96.29 & 73.95 & 57.35 & 49.63 & 33.72 & 50.00 & 27.33 & 55.47     & 1685.14 \\
Qwen2.5-VL-7B + trained-ViT       
& \cellcolor{green!10}96.43 
& \cellcolor{green!10}75.26 
& \cellcolor{green!10}58.75 
& \cellcolor{green!10}50.37 
& \cellcolor{red!5}33.14 
& \cellcolor{green!10}55.00 
& \cellcolor{green!10}31.33 
& \cellcolor{green!10}57.18 
& \cellcolor{green!10}1694.83 \\
Qwen2.5-VL-7B (Ours)        
& \cellcolor{green!10}97.52 
& \cellcolor{green!10}75.97 
& \cellcolor{green!10}59.36 
& \cellcolor{green!10}51.85 
& \cellcolor{green!10}37.79 
& \cellcolor{green!10}59.17 
& \cellcolor{green!10}28.00 
& \cellcolor{green!10}58.52 
& \cellcolor{green!10}1701.87 \\
\bottomrule
\end{tabular}}
\caption{
\textbf{Ablation results.}
\colorbox{green!10}{Green} cells denote an improvement over the corresponding base model, \colorbox{red!5}{red} cells indicate a drop, and \colorbox{yellow!10}{yellow} marks no change.  \emph{All column abbreviations (Pope, Coarse, Fine, Visual\_Sim, Visual\_Corr, Count, MMVP, Ave, MME) carry the same meanings as in Table~\ref{tab:general_bench_results}.}
}
\vspace{-0.7cm}
\label{tab:ablation_results}
\end{table}
\vspace{-0.2cm}
\section{Experiments}
\vspace{-0.2cm}

We evaluate our approach's effectiveness in enhancing fine-grained visual difference comprehension in MLLMs.

In Section \ref{sec:exp:setting}, we outline our experimental settings. Section \ref{sec:exp:med} presents findings from the Micro Edit Detection (MED) benchmark, designed to assess sensitivity to minor yet semantically significant visual differences. Section \ref{sec:exp:other} examines generalization by testing our models across multiple multimodal benchmarks. Section \ref{sec:exp:ablation} offers an ablation study analyzing how our training strategy's key design choices affect performance.

\vspace{-0.2cm}
\subsection{Experimental Settings}
\vspace{-0.2cm}

\label{sec:exp:setting}
To assess our approach's effectiveness, we fine-tune several open-source vision-language models on the Micro Edit Dataset using our SFT loss (Equation~\ref{eq:SFT_loss}). Models include Qwen2-VL-7B \cite{wang2024qwen2}, Qwen2.5-VL-7B \cite{bai2025qwen2}, LLaVA-V1.6-Vicuna-7B \cite{liu2023improved}, and LLaMA-3.2-Vision-Instruct-11B \cite{grattafiori2024llama}. All fine-tuning uses LLaMA-Factory \cite{zheng2024llamafactory} with consistent hyperparameters for fair comparison.

We also evaluate recent commercial closed-source models as performance benchmarks: GPT-4o (2024-08-06) \cite{hurst2024gpt}, GPT-4.1 (2025-04-14) \cite{achiam2023gpt}, Claude 3.7 Sonnet \cite{Claude3S}, Qwen-VL-Plus (2025-03-18) \cite{bai2025qwen2}, and Doubao-1.5-Vision-Pro (2025-03-28). Implementation details are shown in Appendix~\ref{sec:appendix:training_details}.

\vspace{-0.2cm}
\subsection{Experimental Results on MED Benchmark}
\vspace{-0.2cm}
\label{sec:exp:med}
We evaluate our method on the Micro Edit Detection (MED) benchmark, which tests models’ ability to detect and describe minimal yet semantically meaningful visual differences. Table~\ref{tab:med_results} presents results across eleven task categories, comparing human performance, API-based models, and open-source models with and without our fine-tuning.

\textit{Closed-source models still struggle with fine-grained differences understanding compare with human.}
Human annotators achieve 95.21\% accuracy, setting a high bar that current models struggle to reach. Even top closed-source models like GPT-4.1 (54.26\%), Doubao-1.5-vision-pro (51.81\%), and GPT-4o (51.32\%) fall short, particularly in relational tasks like Action, Comparison, and Negation—revealing persistent gaps in fine-grained visual reasoning.

\textit{Our fine-tuning method significantly boosts open-source models.} Our supervised fine-tuning consistently enhances all tested open-source backbones. Qwen2-VL-7B improves from 38.48\% to 47.55\% (+9.07), with major gains in Spatial (+14.28), Action (+6.67), Count (+5.56), and Differentiation (+25.00). Qwen2.5-VL-7B increases from 40.24\% to 51.61\% (+11.37), excelling in Comparison (+21.06), Count (+22.23), and Universality (+14.29). LLaMA-3.2-11B rises from 34.71\% to 40.92\% (+6.21), and LLaVA-V1.6-7B from 40.44\% to 44.04\% (+3.60). These results demonstrate our approach's broad applicability across diverse architectures.

\textit{Our best model outperforms several closed-source models on MED Benchmark.}
Among open-source models, Qwen2.5-VL-7B (Ours) achieves the highest accuracy (51.61\%), rivaling commercial models like GPT-4o (51.32\%) and Doubao-1.5-Vision-Pro (51.81\%). It also surpasses Claude 3.7 Sonnet (40.44\%) and Qwen-VL-Plus (47.36\%), despite its smaller scale and open-source nature. This demonstrates that with fine-grained data and alignment objectives, open-source vision-language models can match closed-source systems in tasks requiring precise visual difference understanding.

\textit{Evaluation on real-world image pairs confirms generalization.}  
We evaluate all models on the MED-Real Set (35 minimally different real-world image pairs; see Section~\ref{sec:data:bench} and Appendix~\ref{sec:appendix:evaluation_med_real_set}) to assess generalization beyond synthetic edits. Fine-tuned models outperform their base versions: Qwen2.5-VL-7B improves from 68.57\% to 74.2\%, LLaVA-V1.6-7B from 34.29\% to 45.71\%, and LLaMA-3.2-11B from 40.00\% to 51.43\%. Qwen2-VL-7B achieves 82.86\% accuracy both pre- and post-fine-tuning, demonstrating strong intrinsic robustness. These results confirm effective generalization to real-world fine-grained visual differences.

\vspace{-0.2cm}
\subsection{Experimental Results on Other Benchmarks} 
\vspace{-0.2cm}
\label{sec:exp:other}

To evaluate generalization beyond the MED benchmark, we test our fine-tuned models on several established multimodal benchmarks, including MMStar, BLINK, POPE, and MME shown in Table~\ref{tab:general_bench_results}.

\textit{Fine-tuning yields consistent and generalizable performance gains.}
As shown in Table \ref{tab:general_bench_results}, across all models, our supervised fine-tuning improves both perception and reasoning. Qwen2.5-VL-7B (Ours) achieves the highest average score (58.52) and MME perception score (1701.87), outperforming all open-source baselines. Notable gains include Count (+9.17), Visual Correspondence (+4.07), and POPE (+1.23), reflecting stronger numeracy, spatial grounding, and hallucination resistance. These improvements hold across model families, confirming the effectiveness of fine-grained supervision.

\textit{Robustness improvements reflected in Hallucination Benchmark.} 
Qwen2.5-VL-7B (Ours) achieves a POPE score of 97.52, ranking among the highest for open-source models, while LLaVA-V1.6-7B (Ours) improves from 95.56 to 97.39. This demonstrates stronger resistance to hallucinated object predictions, even under challenging prompts. Although minor decreases in Fine-grained Perception are observed—likely due to increased sensitivity to small edits—these are outweighed by overall improvements, confirming enhanced precision with minimal trade-offs.

\vspace{-0.3cm}
\subsection{Ablation Study}
\vspace{-0.2cm}

\label{sec:exp:ablation}

To analyze each component's contribution, we perform an ablation study with three settings: (1) base model without fine-tuning, (2) fine-tuning only the visual encoder (ViT) with the language model (LLM) frozen, and (3) joint fine-tuning of ViT and LLM (our full method). Table~\ref{tab:ablation_results} presents the ablation results for Qwen2-VL-7B and Qwen2.5-VL-7B.

\textit{Fine-tuning the vision encoder alone provides solid gains.}
Fine-tuning the vision encoder (ViT) alone yields consistent improvements in perception-heavy tasks. For Qwen2.5-VL-7B, this results in a 1.31 gain in average score and nearly 10 on the MME perception score. Notable gains include Count (+5.00), Coarse Perception (+1.31), and MMVP (+4.00), highlighting that refining visual representations enhances low-level grounding and numeracy.

\textit{Joint fine-tuning further boosts multimodal reasoning.}
Joint fine-tuning of ViT and LLM achieves the best performance across metrics. For Qwen2.5-VL-7B, it improves the average score by +1.34 and MME by +7.04 compared to the vision-only model. Notable gains in Visual Correspondence (+4.65) and POPE (+1.09) underscore the importance of aligning the language head with updated visual features, which reduces hallucinations and enhances cross-modal integration.

\vspace{-0.25cm}
\section{Conclusions and Limitations}
\vspace{-0.25cm}
\label{sec:conclusions_limitations}

We propose a framework to improve fine-grained visual difference understanding in MLLMs by tackling two key challenges: limited semantically controlled data and weak alignment objectives. Our contributions include: (1) a scalable pipeline to generate minimally different image pairs, (2) the Micro Edit Dataset (MED) with 50K samples and a 200-item benchmark, and (3) a fine-tuning strategy with feature-level consistency to enhance robustness to small edits.

\textbf{Limitations and future directions.} Our framework focuses on binary edits between image pairs, with promising directions in extending it to multi-step transformations, temporal reasoning, and compositional edits. Full fine-tuning of vision and language components achieves strong performance, but exploring efficient adaptation methods could reduce computational costs. Applying our approach to unified-token architectures like Chameleon and Emu3 may further enhance its applicability.

\bibliographystyle{ref.bst}  
\small
\bibliography{Reference}
\normalsize

\newpage
\appendix
\appendixpage

\section{Examples of SFT Dataset}
\label{sec:appendix:sft_dataset_examples}
\begin{figure}[H]
    \centering
    \includegraphics[width=1\textwidth]{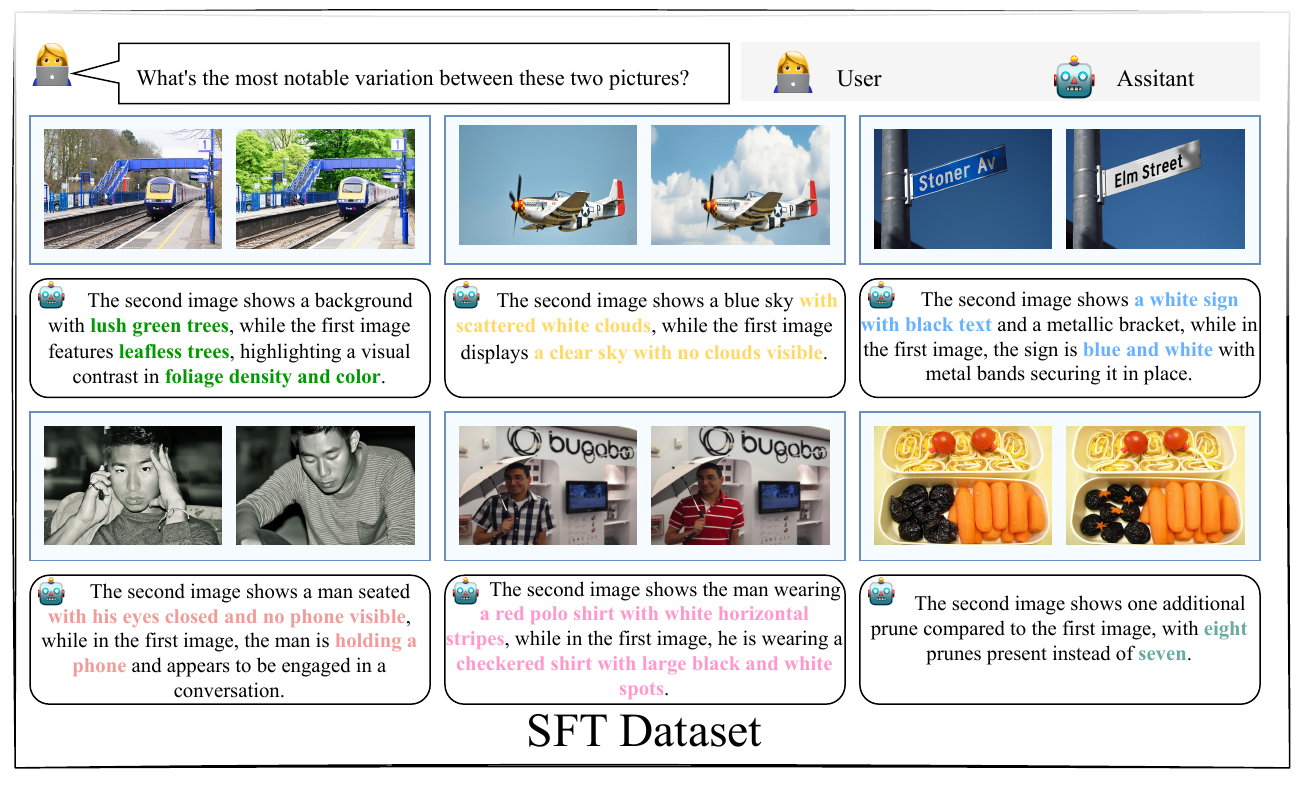}
    \vspace{-0.5cm}
    \caption{\textbf{Example samples from the Micro Edit Dataset (MED) used for supervised fine-tuning.} Each sample consists of a minimally edited image pair, a question prompting for the most notable difference, and a visually grounded answer. Answers are refined to highlight concrete semantic changes based on visual evidence. The full dataset contains over 50K such high-quality QA pairs supporting fine-grained multimodal reasoning.}
    \label{fig:sft_dataset}
    \vspace{-0.5cm}
\end{figure}

In this section, we provide some examples of our constructed MED dataset for supervised fine-tuning in Fig~\ref{fig:sft_dataset}. It combines high-quality image pairs with corresponding questions and visually grounded answers, focusing on fine-grained reasoning tasks.

\section{Experiments on MED-Real Set}
\label{sec:appendix:evaluation_med_real_set}
We provide the evaluation results of the four models on the MED-Real set of our MED Benchmark in Tab~\ref{tab:med_real_set_results}. We evaluate all models on the MED-Real Set, consisting of 35 minimally different real-world image pairs, to assess their generalization beyond synthetic edits. Fine-tuned models show consistent improvements over their base versions: Qwen2.5-VL-7B improves from 68.57\% to 74.29\%, LLaVA-V1.6-7B from 34.29\% to 45.71\%, and LLaMA-3.2-11B from 40.00\% to 51.43\%. Qwen2-VL-7B retains its high accuracy of 82.86\% with or without fine-tuning, demonstrating strong intrinsic robustness. These findings highlight effective generalization to real-world, fine-grained visual differences.
\begin{table}[H]

\small
\centering
\newcolumntype{C}{>{\centering\arraybackslash}p{12mm}}
\begin{tabular}{lC}
\toprule
\textbf{Model} & \textbf{Acc (\%)} \\
\midrule
Qwen2-VL-7B                              & 82.86 \\
Qwen2-VL-7B (Ours)                       & \cellcolor{yellow!10}82.86 \\
Qwen2.5-VL-7B                            & 68.57 \\
Qwen2.5-VL-7B (Ours)                     & \cellcolor{green!10}74.29 \\
LLaVA-V1.6-vicuna-7B               & 34.29 \\
LLaVA-V1.6-vicuna-7B (Ours)        & \cellcolor{green!10}45.71 \\
Llama-3.2-11B-Vision-Instruct         & 40.00 \\
Llama-3.2-11B-Vision-Instruct (Ours)  & \cellcolor{green!10}51.43 \\
\bottomrule
\end{tabular}
\caption{
\textbf{Performance on MED-Real Set.} 
\colorbox{green!10}{Green} cells denote an improvement over the base model; 
\colorbox{yellow!10}{Yellow} cells indicate no change.
}
\label{tab:med_real_set_results}
\end{table}

\section{Training Details}
\label{sec:appendix:training_details}
In this section, we present all the hyperparameters we used to training the three kinds of models in Table~\ref{appendix:training_details_qwen}, Table~\ref{appendix:training_details_llava} and Table~\ref{appendix:training_details_llama}. All the training processes were conducted using llamafactory \cite{zheng2024llamafactory}. Regarding image resolution and the number of image tokens, we adhere to the original settings specified by each model.
\begin{table}[H]
  \centering
  \small

  \newcolumntype{C}{>{\centering\arraybackslash}p{9mm}}
  \caption{\textbf{Hyperparameters for training Qwen2-VL \& Qwen2.5-VL models}}
  \label{tab:hyperparams}
  \begin{tabular}{@{}l l@{}}
    \toprule
    \textbf{Hyperparameter}                     & \textbf{Value}                   \\
    \midrule
    LoRA Rank                                   & 8                              \\
    LoRA $\alpha$                               & 16                              \\
    LoRA Dropout                                & 0.1                              \\
    LoRA Target                                 & all                              \\
    GPU                                         & 8 $\times$ NVIDIA A800           \\
    Batch Size                                  & 16                                \\
    Gradient Accumulation Steps                 & 8                                \\
    Warmup Ratio                                & 0.1                             \\
    Learning Rate                               & 1e-4                            \\
    Learning Rate Scheduler                     & Cosine                           \\
    Unfreeze Vision Tower                       & True                           \\
    \bottomrule
  \end{tabular}
  \label{appendix:training_details_qwen}
\end{table}

\begin{table}[H]
  \centering
  \small

  \newcolumntype{C}{>{\centering\arraybackslash}p{9mm}}
  \caption{\textbf{Hyperparameters for training LLaVA-V1.6-7B model}}
  \label{tab:hyperparams}
  \begin{tabular}{@{}l l@{}}
    \toprule
    \textbf{Hyperparameter}                     & \textbf{Value}                   \\
    \midrule
    LoRA Rank                                   & 8                              \\
    LoRA $\alpha$                               & 16                              \\
    LoRA Dropout                                & 0.1                              \\
    LoRA Target                                 & all                              \\
    GPU                                         & 8 $\times$ NVIDIA A800           \\
    Batch Size                                  & 1                                \\
    Gradient Accumulation Steps                 & 8                                \\
    Warmup Ratio                                & 0.1                             \\
    Learning Rate                               & 1e-5                            \\
    Learning Rate Scheduler                     & Cosine                           \\
    Unfreeze Vision Tower                       & False                           \\
    \bottomrule
  \end{tabular}
  \label{appendix:training_details_llava}
\end{table}

\begin{table}[H]
  \centering
  \small
  \newcolumntype{C}{>{\centering\arraybackslash}p{9mm}}
  \caption{\textbf{Hyperparameters for training LLaMA-3.2-11B model}}
  \label{tab:hyperparams}
  \begin{tabular}{@{}l l@{}}
    \toprule
    \textbf{Hyperparameter}                     & \textbf{Value}                   \\
    \midrule
    LoRA Rank                                   & 8                              \\
    LoRA $\alpha$                               & 16                              \\
    LoRA Dropout                                & 0.1                              \\
    LoRA Target                                 & all                              \\
    GPU                                         & 8 $\times$ NVIDIA A800           \\
    Batch Size                                  & 4                                \\
    Gradient Accumulation Steps                 & 4                                \\
    Warmup Ratio                                & 0.1                             \\
    Learning Rate                               & 1e-5                            \\
    Learning Rate Scheduler                     & Cosine                           \\
    Unfreeze Vision Tower                       & True                           \\
    \bottomrule
  \end{tabular}
  \label{appendix:training_details_llama}
\end{table}
\section{More Examples of MED Benchmark}
\label{sec:appendix:benchmark_examples}
\begin{figure}[H]
    \centering
    \includegraphics[width=\textwidth]{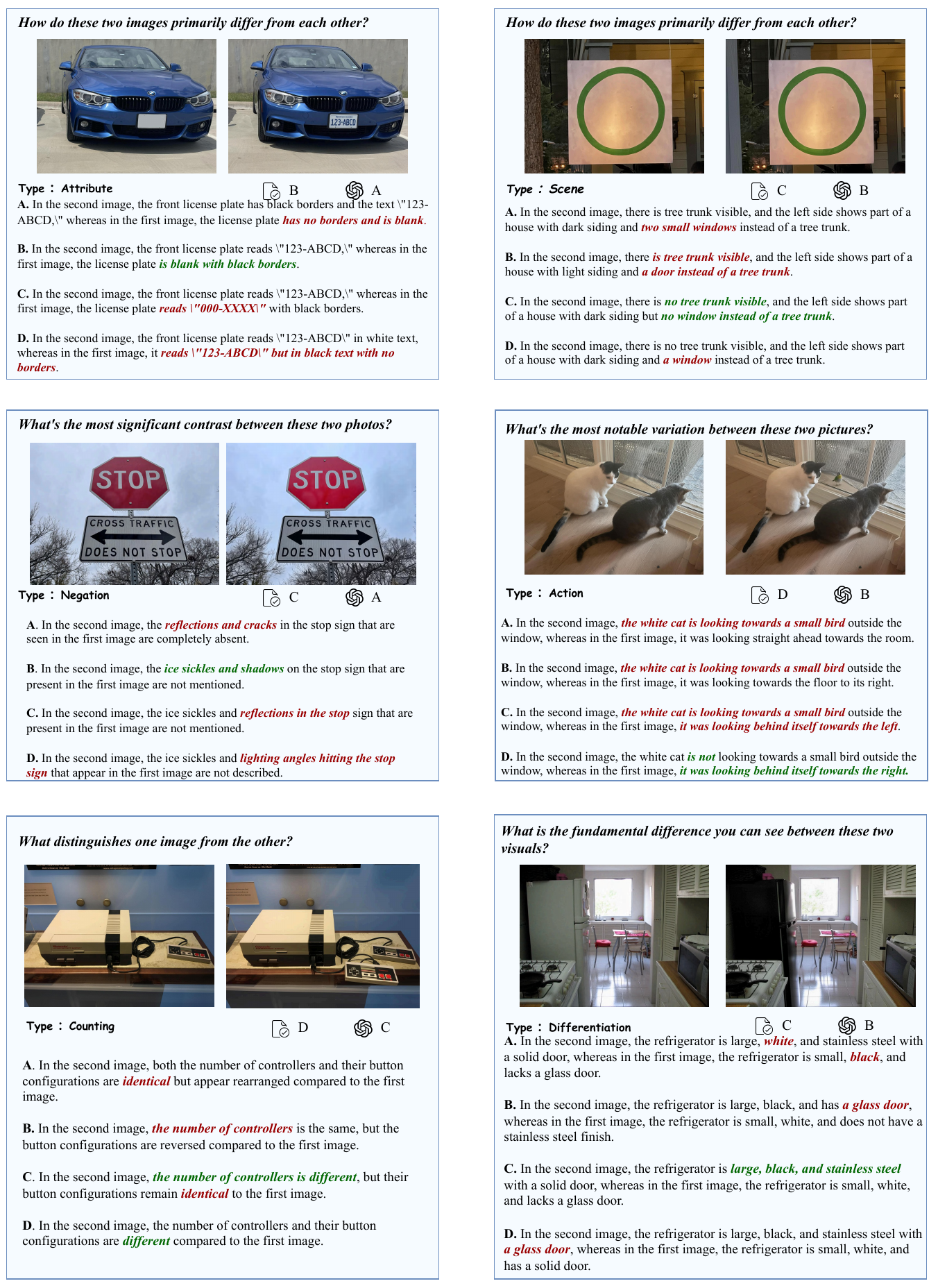}
    \caption{More examples in the MED benchmark}
    \label{fig:Benchmark_more_1}
\end{figure}

\begin{figure}[H]
    \centering
    \includegraphics[width=\textwidth]{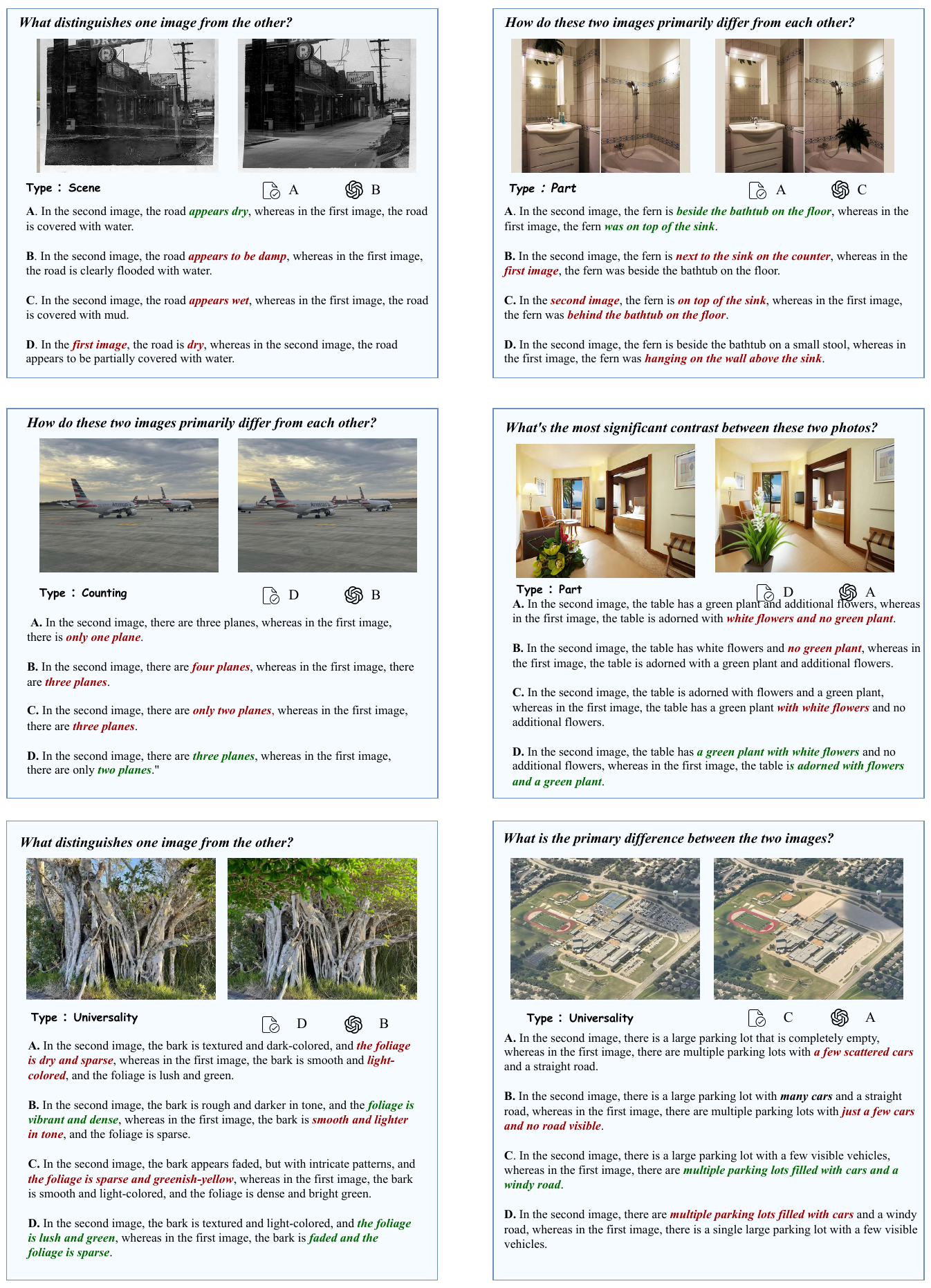}
    \caption{More examples in the MED benchmark}
    \label{fig:Benchmark_more_2}
\end{figure}

In this section, we present additional examples from the MED Benchmark in Fig~\ref{fig:Benchmark_more_1} and Fig~\ref{fig:Benchmark_more_2}, including their types, answer options, correct answers, and the answers provided by GPT-4.1.

\clearpage

\section{Prompts}
\label{sec:appendix:prompt}
In this section, we systematically present all the prompts used in the construction of the Micro Edit Dataset. These prompts are designed to cover every stage of the dataset creation process, including filtering editable images, generating editing instructions, producing detailed descriptions for both original and edited images, describing the differences between image pairs, conducting evaluation, and generating training as well as evaluation data.

\begin{figure}[H]
    \centering
    \includegraphics[width=1\textwidth]{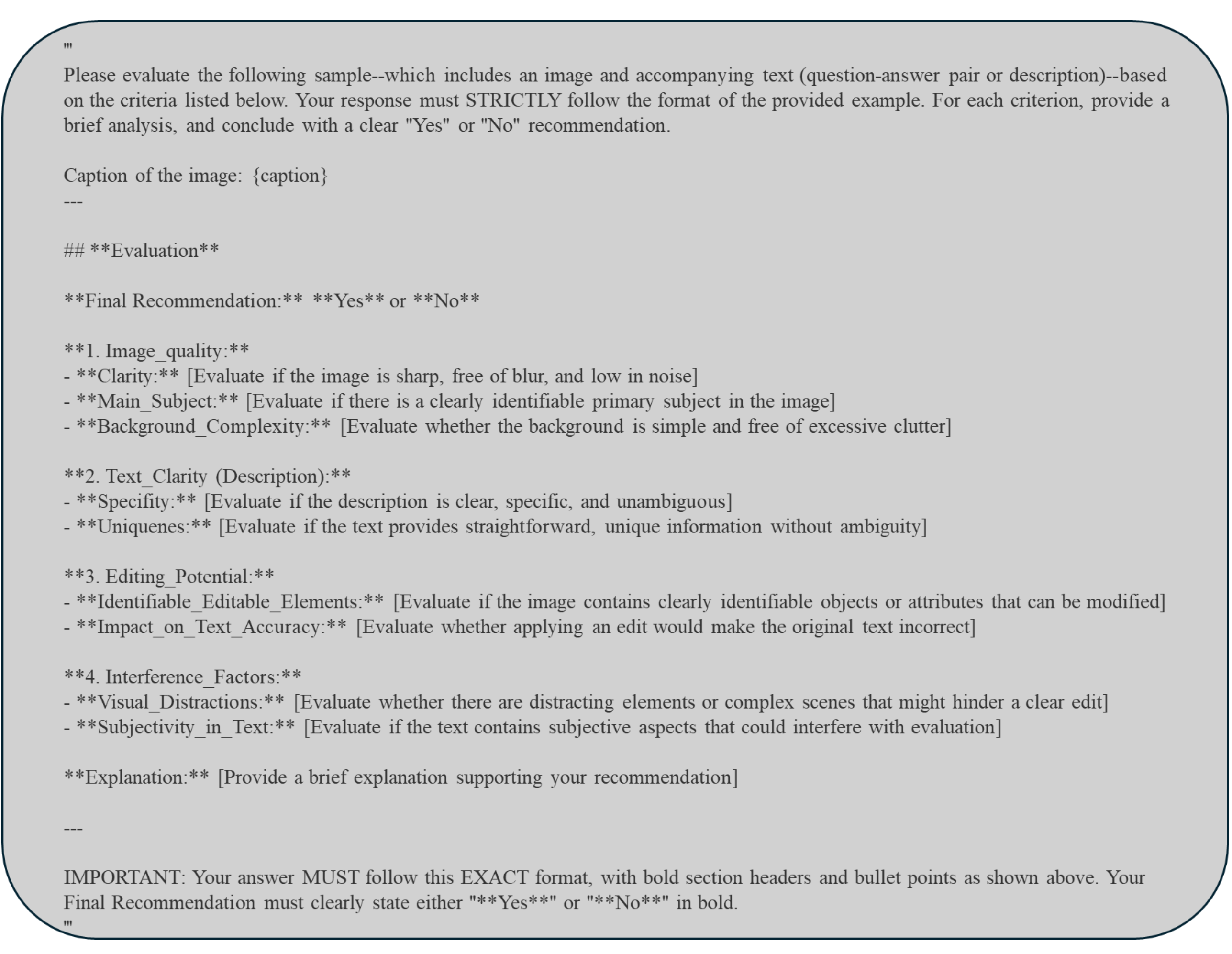}
    \vspace{-0.5cm}
    \caption{Prompt for Qwen-2.5-VL-72B to filter editable images}
    \label{fig:filter_edit_prompt}
    \vspace{-0.5cm}  
\end{figure}

\begin{figure}[H]
    \centering
    \includegraphics[width=1\textwidth]{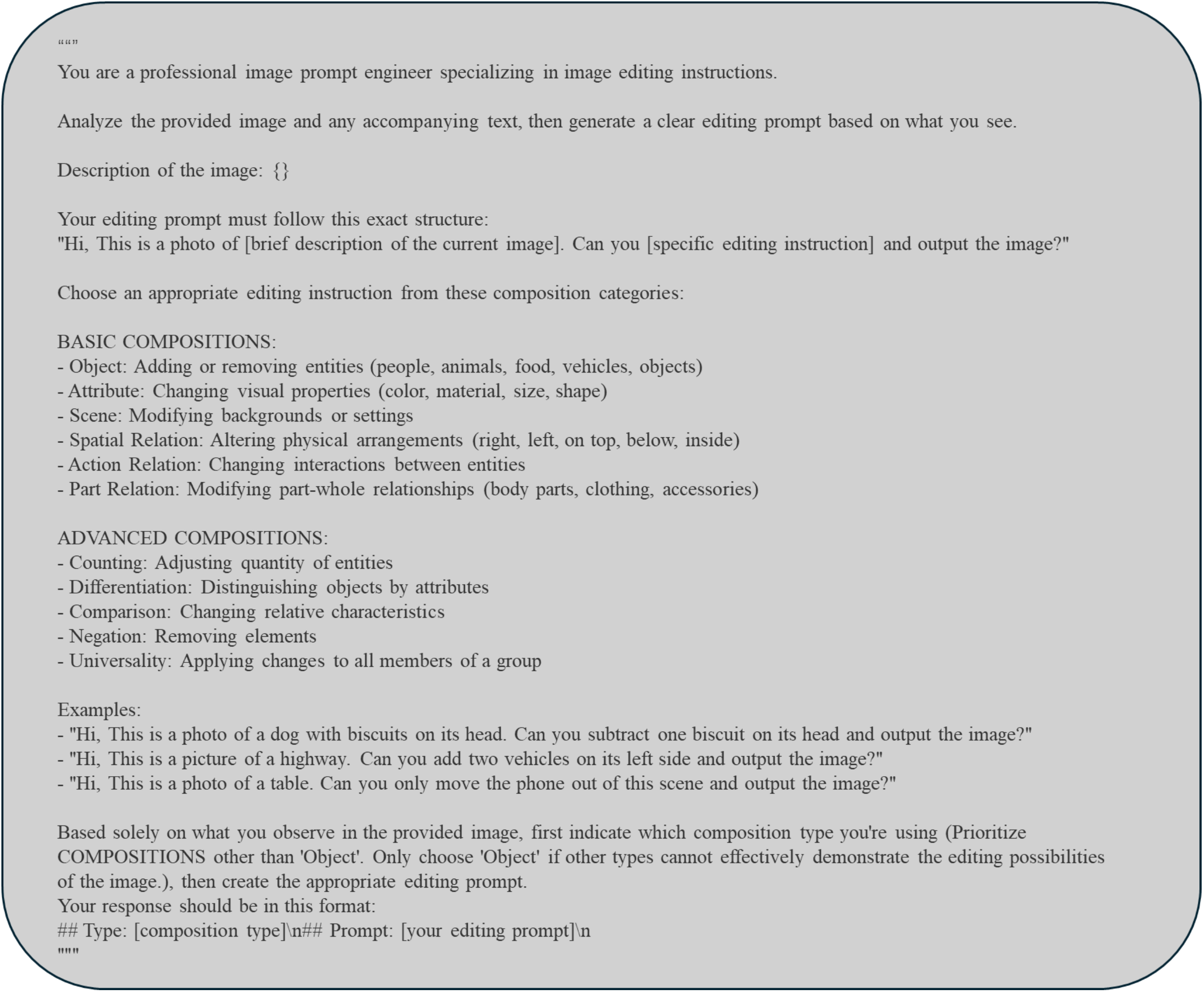}
    \vspace{-0.5cm}
    \caption{Prompt for Qwen-2.5-VL-72B to generate editing prompt}
    \label{fig:edit_prompt}
    \vspace{-0.5cm}  
\end{figure}

\begin{figure}[H]
    \centering
    \includegraphics[width=1\textwidth]{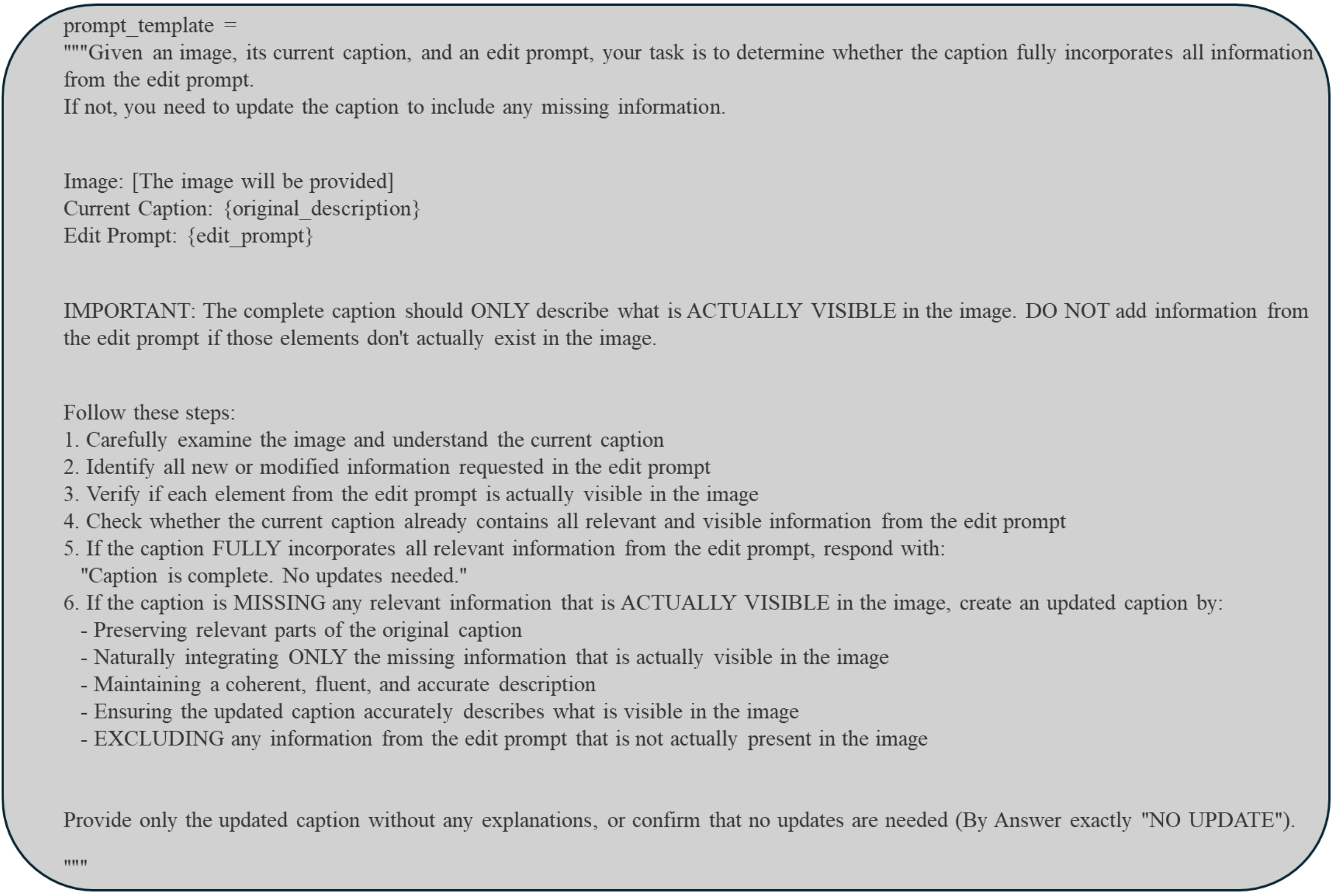}
    \vspace{-0.5cm}
    \caption{Prompt for Qwen-2.5-VL-72B to generate complete original description for original images}
    \label{fig:Original_Description}
    \vspace{-0.5cm}  
\end{figure}

\begin{figure}[H]
    \centering
    \includegraphics[width=1\textwidth]{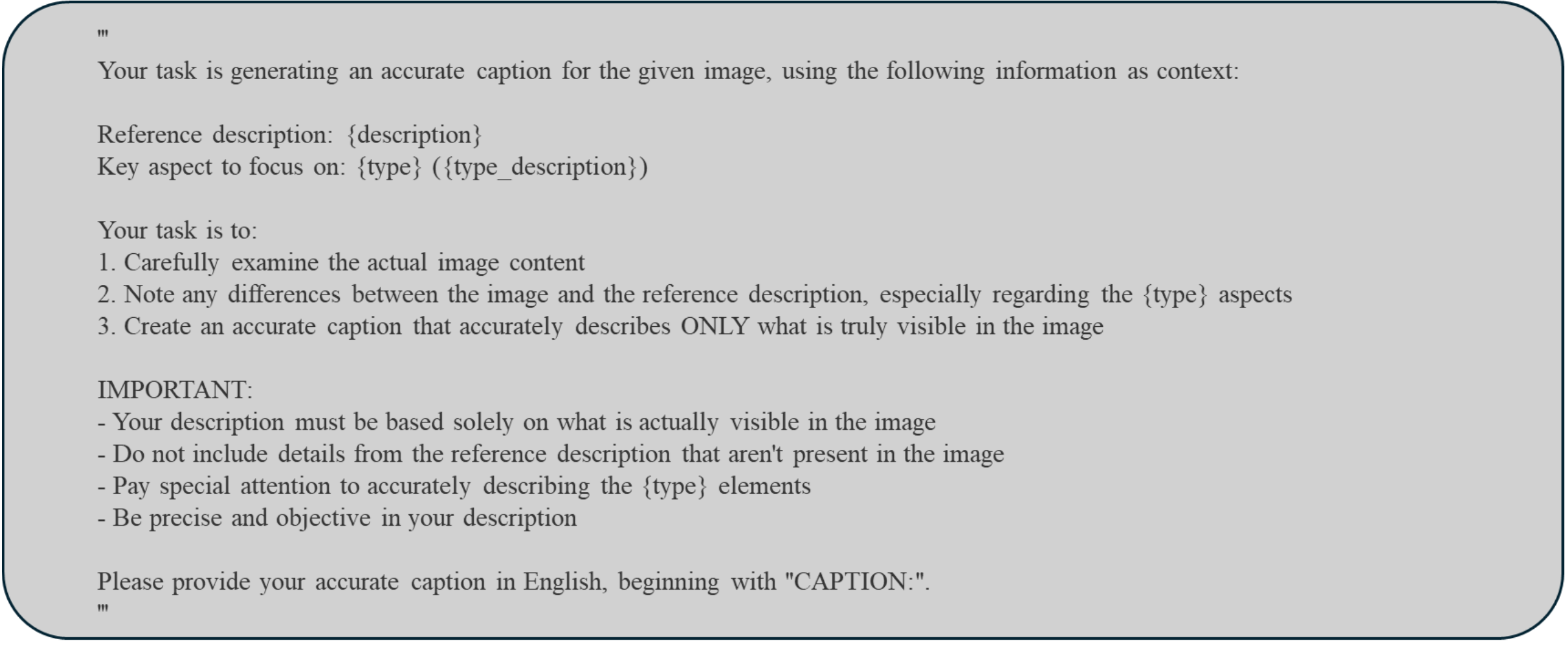}
    \vspace{-0.5cm}
    \caption{Prompt for Qwen-2.5-VL-72B to generate description for edited images}
    \label{fig:edit_Description}
    \vspace{-0.5cm}  
\end{figure}

\begin{figure}[H]
    \centering
    \includegraphics[width=1\textwidth]{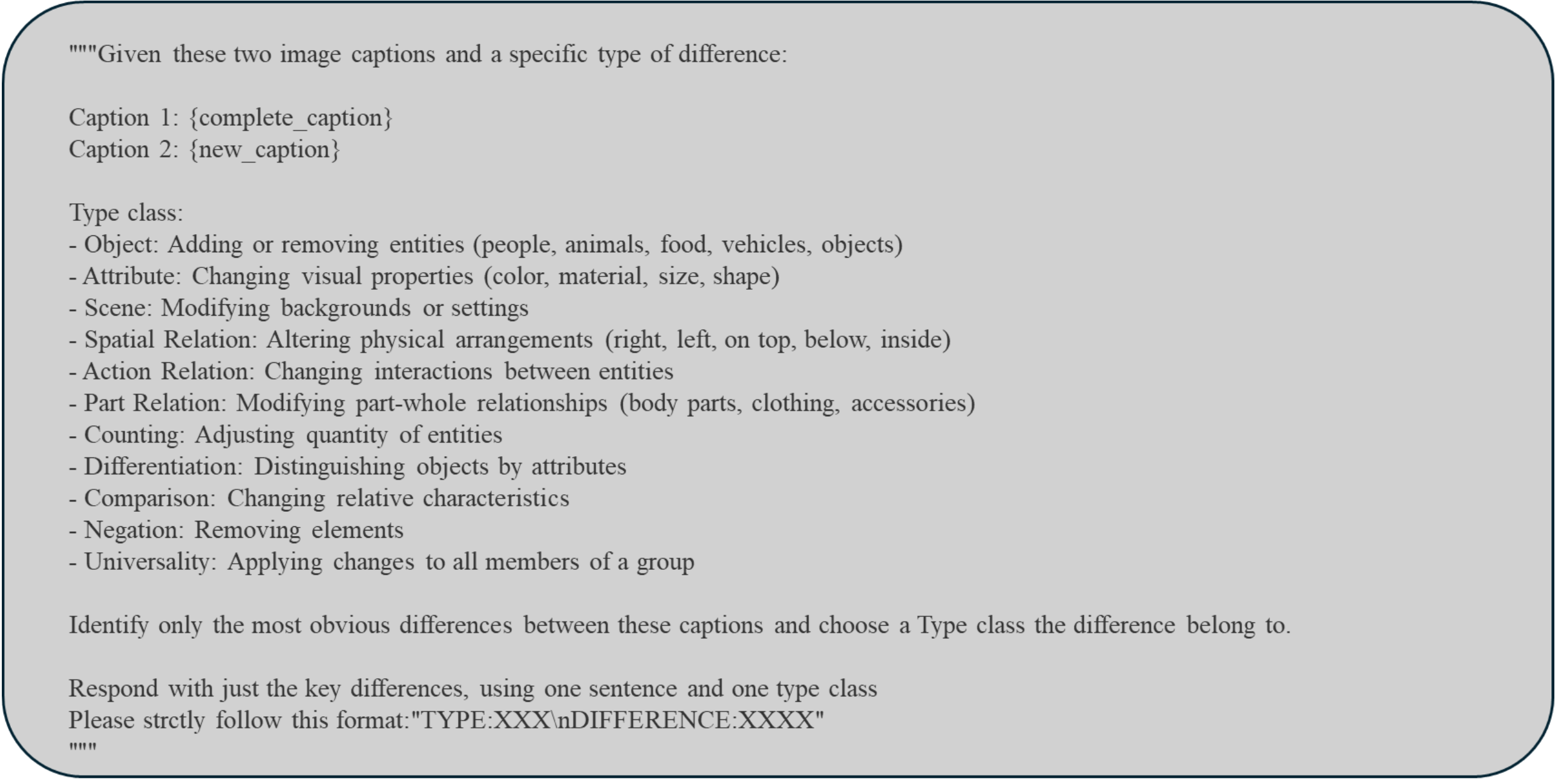}
    \vspace{-0.5cm}
    \caption{Prompt for Qwen-2.5-VL-72B to generate difference description}
    \label{fig:_Difference_Description}
    \vspace{-0.5cm}  
\end{figure}

\begin{figure}[H]
    \centering
    \includegraphics[width=1\textwidth]{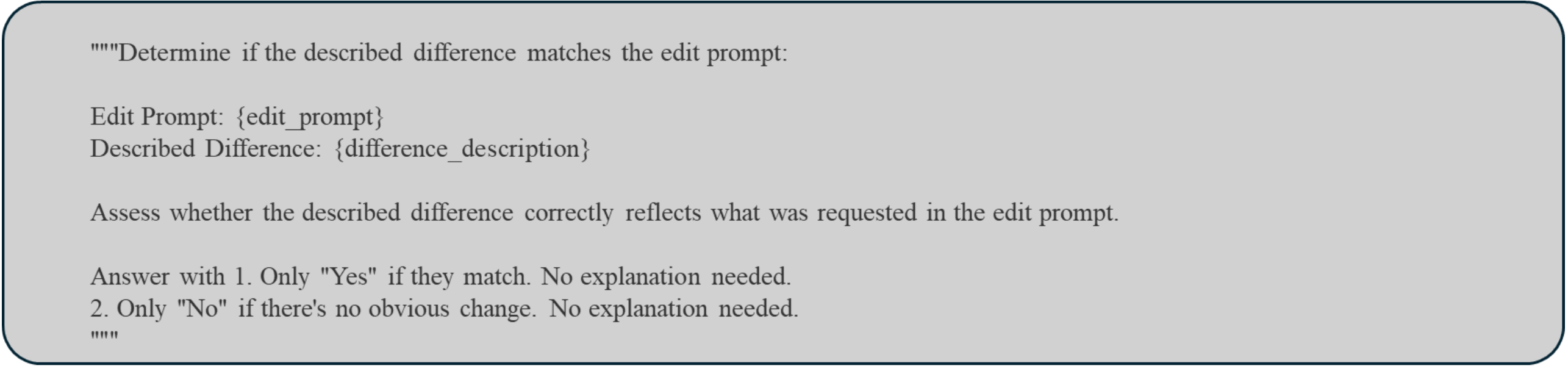}
    \vspace{-0.5cm}
    \caption{Prompt for Qwen-2.5-VL-72B to get judge}
    \label{fig:get_judge}
    \vspace{-0.5cm}  
\end{figure}

\begin{figure}[H]
    \centering
    \includegraphics[width=1\textwidth]{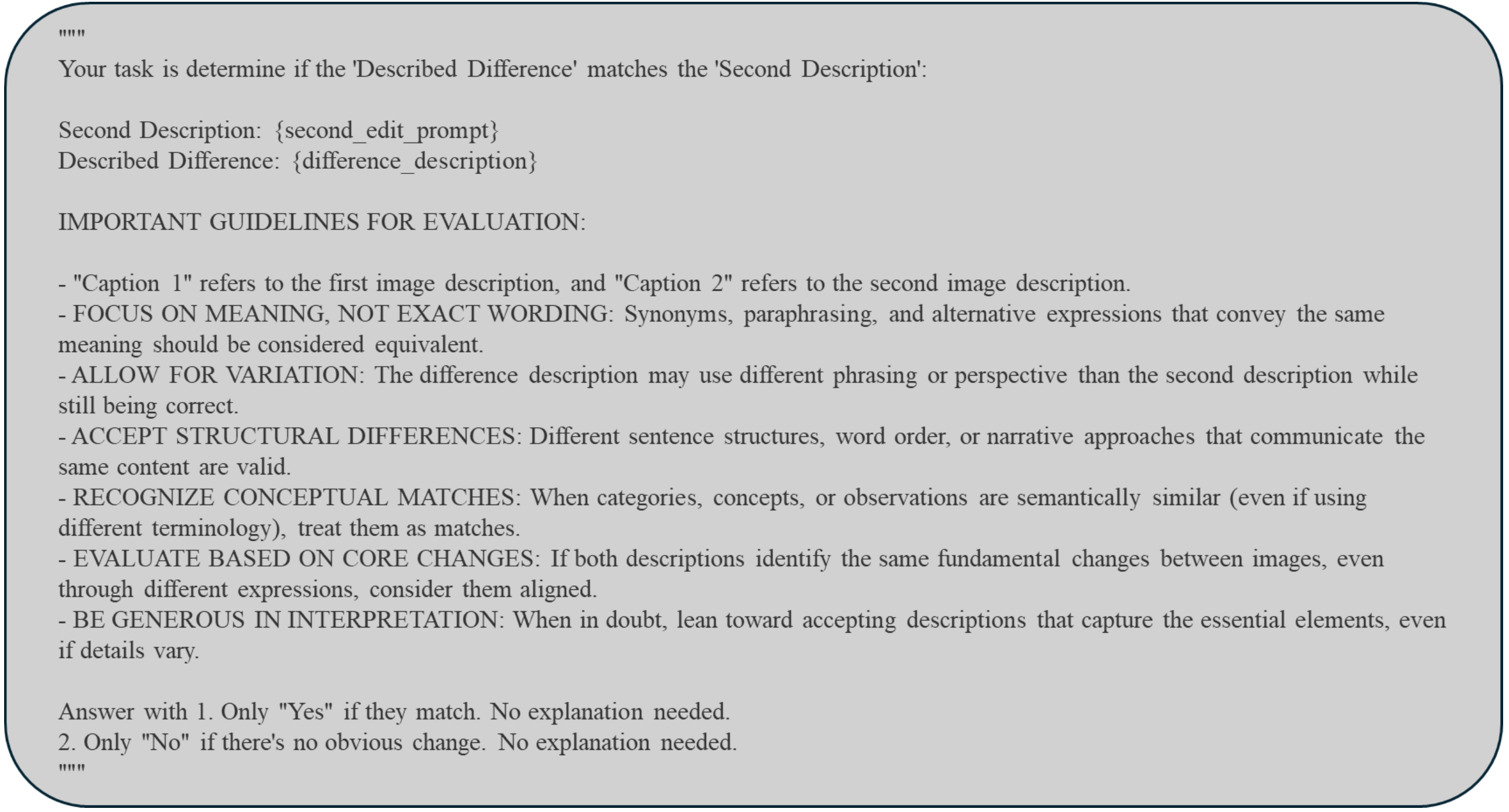}
    \vspace{-0.5cm}
    \caption{Prompt for Qwen-2.5-VL-72B to get second judge}
    \label{fig:Get_Second_Judge}
    \vspace{-0.5cm}  
\end{figure}

\begin{figure}[H]
    \centering
    \includegraphics[width=1\textwidth]{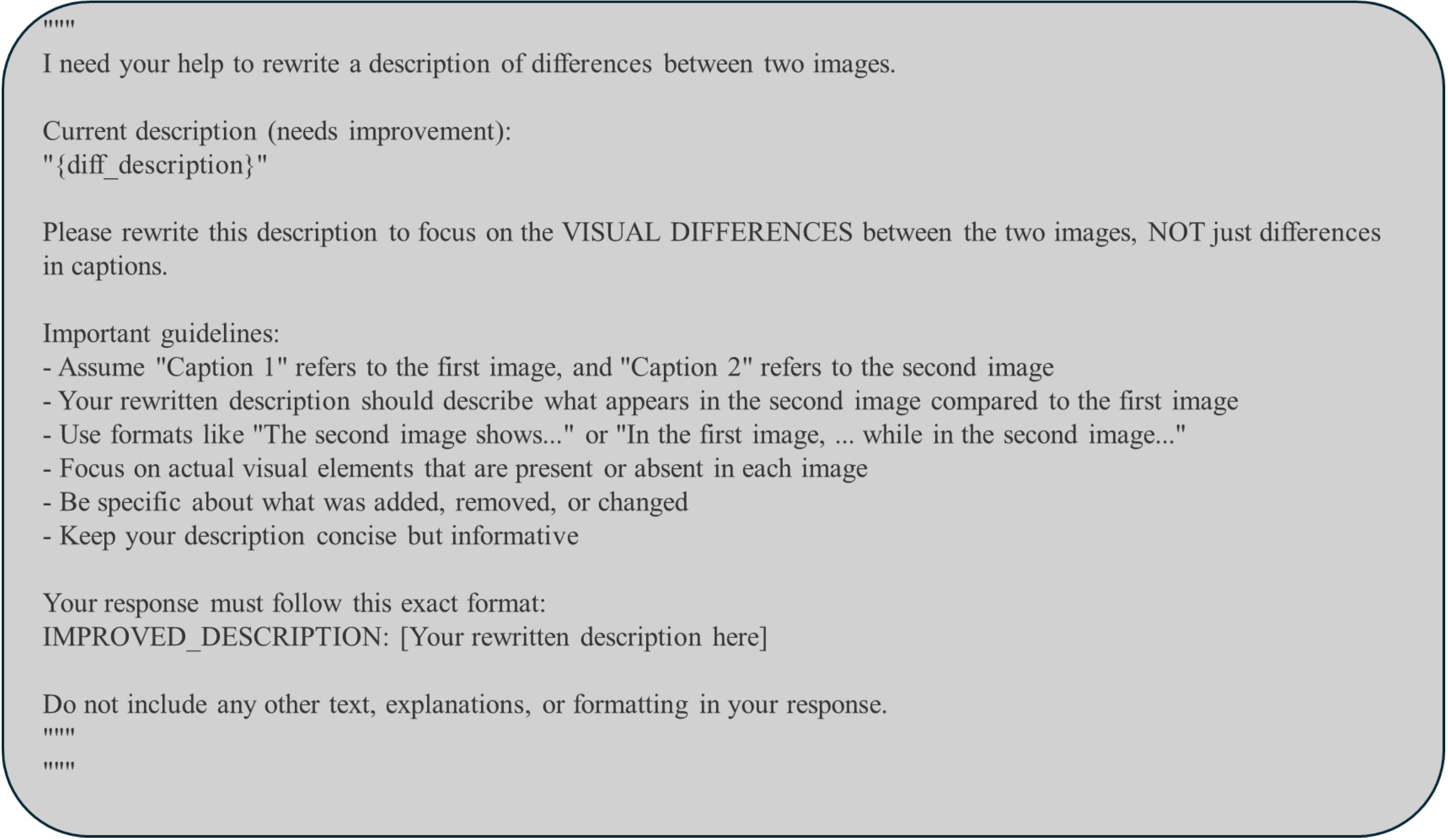}
    \vspace{-0.5cm}
    \caption{Prompt for Qwen3-32B to generate SFT data}
    \label{fig:Get_SFT_Data}
    \vspace{-0.5cm}  
\end{figure}

\begin{figure}[H]
    \centering
    \includegraphics[width=1\textwidth]{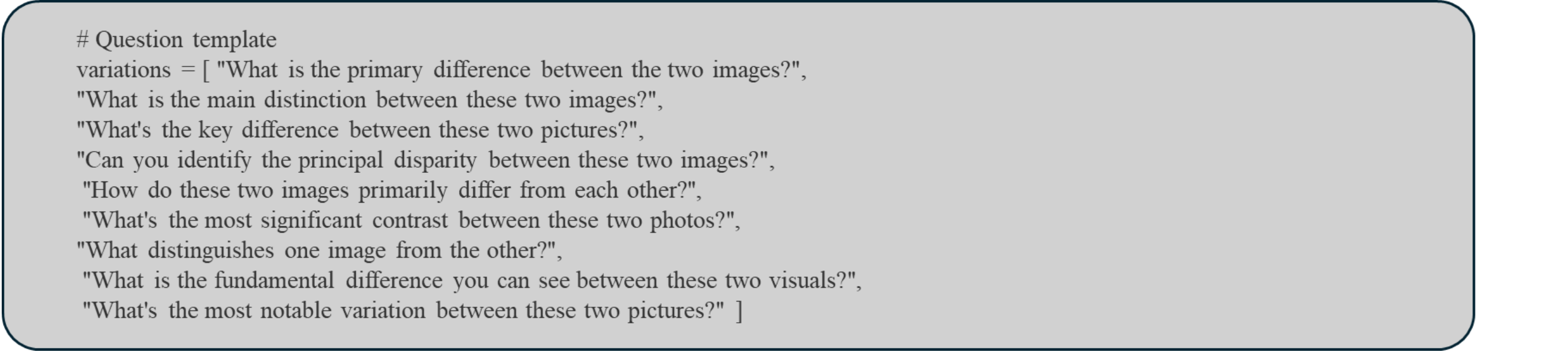}
    \vspace{-0.5cm}
    \caption{Question Template of Benchmark}
    \label{fig:question}
    \vspace{-0.5cm}  
\end{figure}

\begin{figure}[H]
    \centering
    \includegraphics[width=1\textwidth]{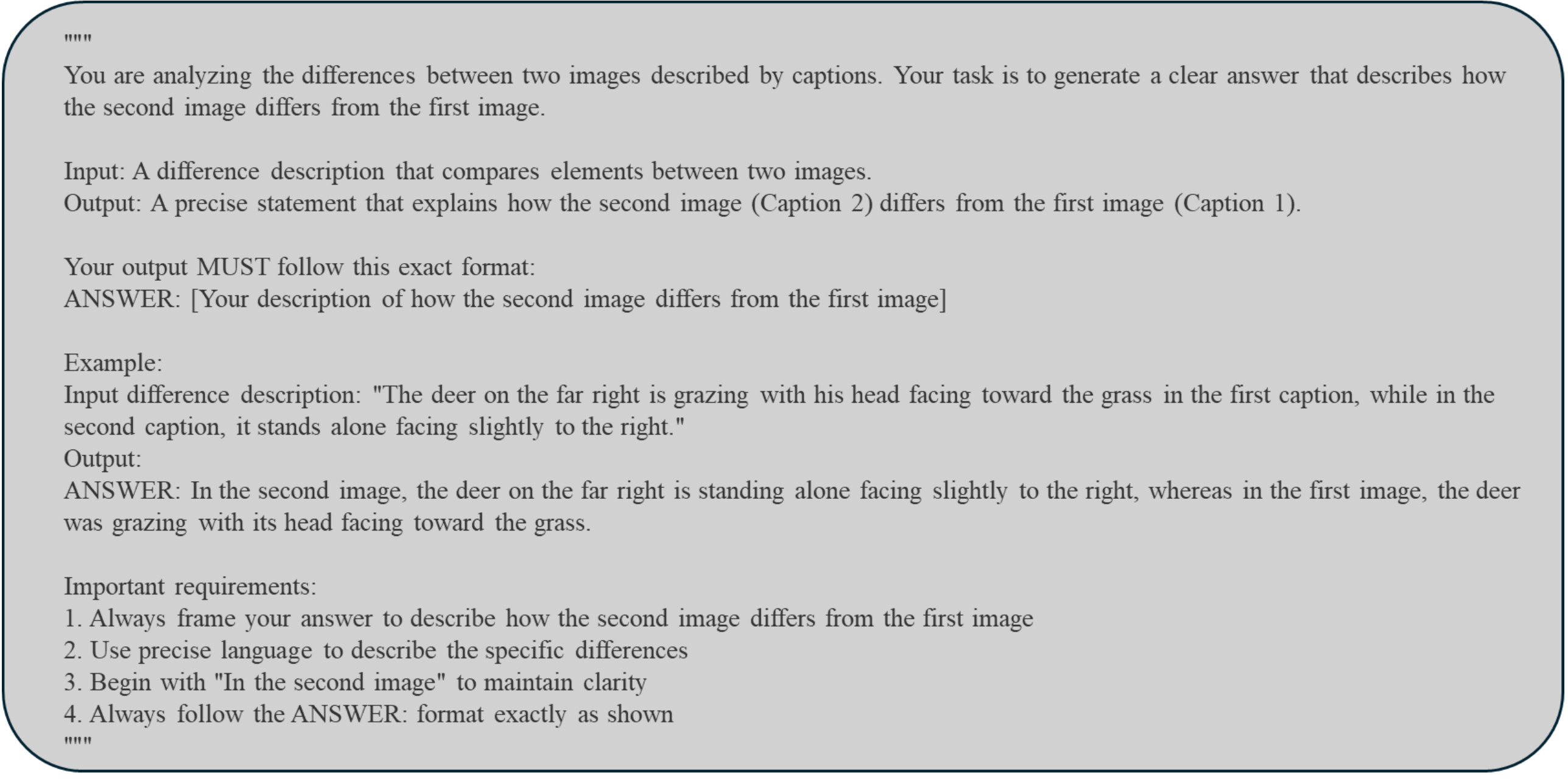}
    \vspace{-0.5cm}
    \caption{Prompt for GPT-4o to generate right answer}
    \label{fig:Generating_right_answer}
    \vspace{-0.5cm}  
\end{figure}

\begin{figure}[H]
    \centering
    \includegraphics[width=1\textwidth]{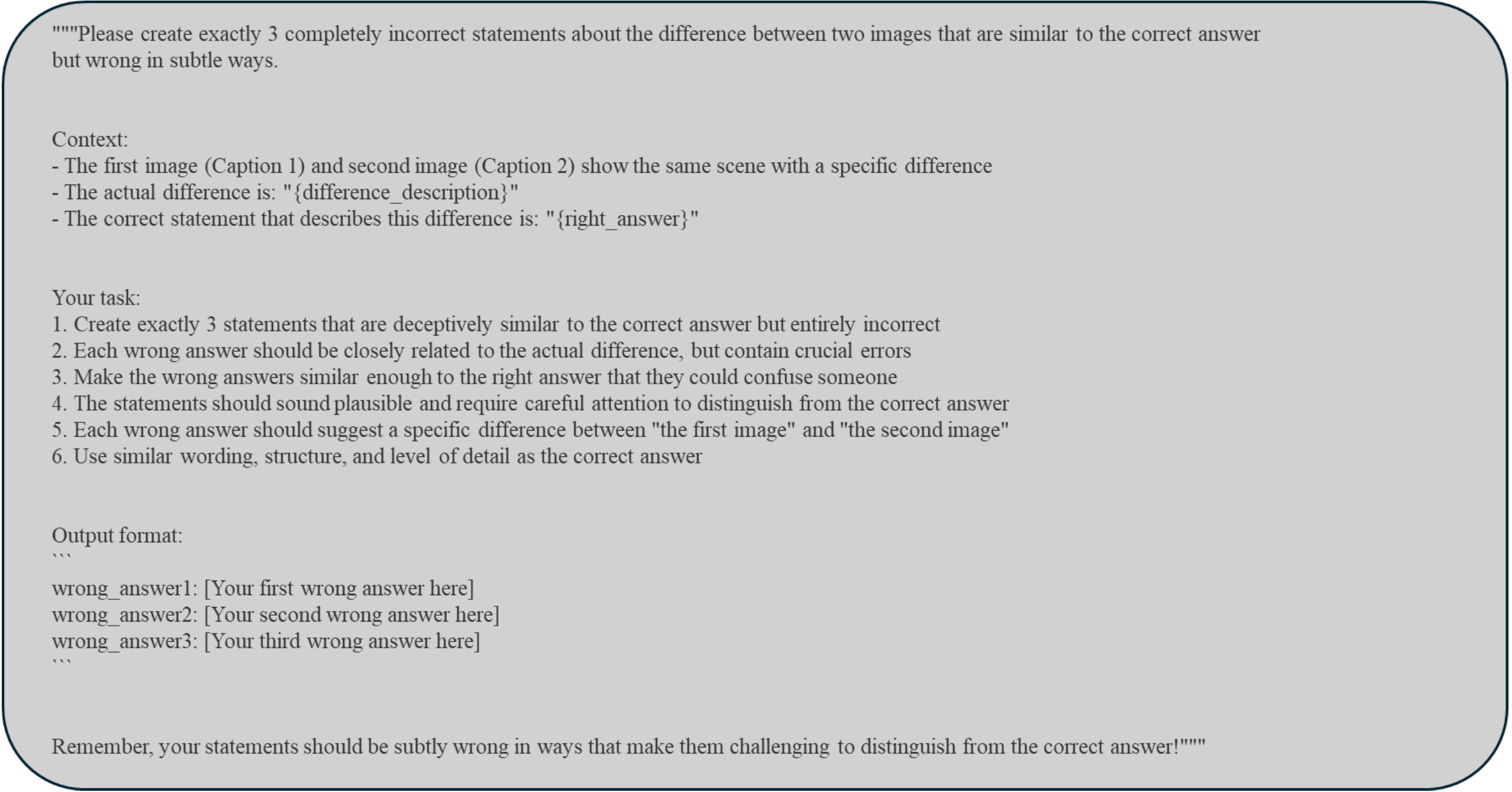}
    \vspace{-0.5cm}
    \caption{Prompt for GPT-4o to generate wrong answer}
    \label{fig:Generating_wrong_answer}
    \vspace{-0.5cm}  
\end{figure}

Specifically:
\begin{itemize}
  \item Figure~\ref{fig:filter_edit_prompt} shows the prompt for filtering editable images, ensuring that the selected images are suitable for subsequent editing tasks.
  \item Figure~\ref{fig:edit_prompt} illustrates the prompt for generating image editing instructions, which enables the creation of diverse editing scenarios.
  \item Figures~\ref{fig:Original_Description} and~\ref{fig:edit_Description} demonstrate the prompts for generating comprehensive descriptions of the original and the edited images, respectively, providing essential context for later analysis.
  \item Figure~\ref{fig:_Difference_Description} presents the prompt for describing the differences between two images, facilitating both quantitative and qualitative analysis of micro edits.
  \item Figures~\ref{fig:get_judge} and~\ref{fig:Get_Second_Judge} display the prompts used for evaluating the quality of the image edits, ensuring an objective assessment process.
  \item Figure~\ref{fig:Get_SFT_Data} shows the prompt for generating Supervised Fine-Tuning (SFT) data, which supplies high-quality training samples for model fine-tuning.
  \item Figure~\ref{fig:question} provides the question template used for benchmark evaluations, enabling a systematic assessment of model performance on micro editing tasks.
  \item Figures~\ref{fig:Generating_right_answer} and~\ref{fig:Generating_wrong_answer} demonstrate the prompts for generating the correct and distractor answers, respectively, enhancing the scientific rigor and difficulty of the benchmark.
\end{itemize}

All prompts have been carefully crafted and iteratively refined to ensure high quality and diversity in the dataset, laying a solid foundation for subsequent research and algorithm development on micro editing tasks.

\end{document}